\RequirePackage{fix-cm}
\documentclass{svjour3}                     		
\smartqed  
\usepackage{authblk}
\usepackage{color}
\usepackage{marvosym}

\usepackage{graphicx}

\usepackage{multirow}
\usepackage{color, colortbl}
\definecolor{Gray}{gray}{0.8}

\usepackage{float}
\usepackage{subfigure}
\usepackage[percent]{overpic}
\usepackage{caption}
\usepackage{algorithm}
\usepackage{algpseudocode}

\makeatletter
\renewcommand{\ALG@beginalgorithmic}
\makeatother

\usepackage{siunitx}
\usepackage{amsmath}
\usepackage{amssymb}
\usepackage{cite}
\usepackage{url}
\usepackage{textcomp}
\usepackage[utf8]{inputenc}
\errorcontextlines\maxdimen

\makeatletter
\newcommand*{\algrule}[1][\algorithmicindent]{\makebox[#1][l]{\hspace*{.5em}\thealgruleextra\vrule height \thealgruleheight depth \thealgruledepth}}%
\newcommand*{\thealgruleextra}{}
\newcommand*{\thealgruleheight}{1\baselineskip}
\newcommand*{\thealgruledepth}{.3\baselineskip}

\newcount\ALG@printindent@tempcnta
\def\ALG@printindent{%
    \ifnum \theALG@nested>0
        \ifx\ALG@text\ALG@x@notext
        \else
            \unskip
            \addvspace{-1pt}
            \ALG@printindent@tempcnta=1
            \loop
                \algrule[\csname ALG@ind@\the\ALG@printindent@tempcnta\endcsname]%
                \advance \ALG@printindent@tempcnta 1
            \ifnum \ALG@printindent@tempcnta<\numexpr\theALG@nested+1\relax
            \repeat
        \fi
    \fi
    }%
\usepackage{etoolbox}
\patchcmd{\ALG@doentity}{\noindent\hskip\ALG@tlm}{\ALG@printindent}{}{\errmessage{failed to patch}}
\makeatother

\newbox\statebox
\newcommand{\myState}[1]{%
    \setbox\statebox=\vbox{#1}%
    \edef\thealgruleheight{\dimexpr \the\ht\statebox+1pt\relax}%
    \edef\thealgruledepth{\dimexpr \the\dp\statebox+1pt\relax}%
    \ifdim\thealgruleheight<.75\baselineskip
        \def\thealgruleheight{\dimexpr .75\baselineskip+1pt\relax}%
    \fi
    \ifdim\thealgruledepth<.25\baselineskip
        \def\thealgruledepth{\dimexpr .25\baselineskip+1pt\relax}%
    \fi
    \State #1%
    \def\thealgruleheight{\dimexpr .75\baselineskip+1pt\relax}%
    \def\thealgruledepth{\dimexpr .25\baselineskip+1pt\relax}%
}
\newcommand\longvdots[1]{\raisebox{1em}{\rotatebox{-90}{\hbox to #1 {\dotfill}}}}              
\usepackage{mathptmx}      
\title{Detection, localisation and tracking of pallets using machine learning techniques and 2D range data
\thanks{The research leading to these results has received funding from the POR/FESR Liguria regional funding scheme, under grant agreement n. 56 (AIRONE).}
}

\titlerunning{Detection, localisation and tracking of pallets}  
      
\author{Ihab S. Mohamed
   \and Alessio Capitanelli
   \and Fulvio Mastrogiovanni
   \and Stefano Rovetta
   \and Renato Zaccaria}

\authorrunning{Ihab S. Mohamed et al.} 

\institute{
Ihab S. Mohamed \Letter
\at Department of Informatics, Bioengineering, Robotics and Systems Engineering, University of Genoa, Italy \\
\email{imohamed@mion.elka.pw.edu.pl}\\
Tel.: +2011-286-10-293
\and
Alessio Capitanelli
\at Department of Informatics, Bioengineering, Robotics and Systems Engineering, University of Genoa, Italy \\
\email{alessio.capitanelli@dibris.unige.it} \\
\and
Fulvio Mastrogiovanni
\at Department of Informatics, Bioengineering, Robotics and Systems Engineering, University of Genoa, Italy \\
\email{fulvio.mastrogiovanni@unige.it} \\
\and
Stefano Rovetta
\at Department of Informatics, Bioengineering, Robotics and Systems Engineering, University of Genoa, Italy \\
\email{stefano.rovetta@unige.it} \\
\and
Renato Zaccaria
\at Department of Informatics, Bioengineering, Robotics and Systems Engineering, University of Genoa, Italy \\
\email{renato.zaccaria@unige.it} \\
}

\date{Received: date / Accepted: date}

\begin{document}

\maketitle

\begin{abstract}

The problem of autonomous transportation in industrial scenarios is receiving a renewed interest due to the way it can revolutionise internal logistics, especially in unstructured environments.
This paper presents a novel architecture allowing a robot to detect, localise, and track (possibly multiple) pallets using machine learning techniques based on an on-board 2D laser rangefinder only.
The architecture is composed of two main components: 
the first stage is a pallet detector employing a Faster Region-based Convolutional Neural Network (Faster R-CNN) detector cascaded with a CNN-based classifier;
the second stage is a Kalman filter for localising and tracking detected pallets, which we also use to defer commitment to a pallet detected in the first stage until sufficient confidence has been acquired via a sequential data acquisition process.
For fine-tuning the CNNs, the architecture has been systematically evaluated using a real-world dataset containing 340 labeled 2D scans, which have been made freely available in an online repository.
Detection performance has been assessed on the basis of the average accuracy over k-fold cross-validation, and it scored 99.58\% in our tests.
Concerning pallet localisation and tracking, experiments have been performed in a scenario where the robot is approaching the pallet to fork. 
Although data have been originally acquired by considering only one pallet as per specification of the use case we consider, artificial data have been generated as well to mimic the presence of multiple pallets in the robot workspace. 
Our experimental results confirm that the system is capable of identifying, localising and tracking pallets with a high success rate while being robust to false positives.


\end{abstract}

\section{Introduction}
\label{sec:intro}

The adoption of the Industry 4.0 paradigm is thought to intrinsically change the nature of shop-floor and warehouse environments along many dimensions, and the use of autonomous mobile robots for inbound freight transportation and delivery is no exception \cite{DAndrea2012}.
Traditionally, automated guided vehicles (AGVs) have been adopted in industrial environments for freight transportation and delivery under a number of assumptions, namely:
\begin{enumerate} 
\item a well-defined, structured, and \textit{obstacle free} workspace for robot navigation, and
\item unambiguous robot sensing and perception capabilities as far as their interaction with the environment is concerned.
\end{enumerate}    
Nowadays, in spite of high levels in shop-floor and warehouse automation, such assumptions largely still hold, even in case of novel solutions proposed by the start-up ecosystem, with a few notable exceptions such as the one commercialised by Otto Motors\footnote{Web: \url{www.ottomotors.com}.} and Fetch Robotics\footnote{Web: \url{fetchrobotics.com/}.}.
However, the tenets of the Industry 4.0 paradigm are expected to require relaxing such assumptions.
Given the goal of providing customers with personalised and just-in-time delivery of products, it is foreseen that warehouse environments will become more \textit{dynamic} and human-friendly, and will host human-robot collaborative processes to a great extent \cite{Krugeretal2009, Heyer2010, Darvishetal2018}.
As far as AGVs are concerned, such directives imply higher standards in autonomy, as well as more robust perception and decision making capabilities.

Notwithstanding such ferment, pallets are still considered of the utmost importance in warehouses.
According to a survey by Peerless Research Group\footnote{Web: \url{www.peerlessresearch.com/}.}, pallets are preferred over novel automated logistics systems for a number of reasons: purchase price ($60.9\%$ of the qualified responses), strength ($55.6\%$), durability ($54.3\%$), and reusability ($44.4\%$), just to name a few.  
Among the various materials employed for pallet design and manufacturing, wood pallets are the preferred ones.
When asked how many pallets survey respondents are using with respect to what they did one year before, $46\%$ of them declare using approximately the same number of pallets, $42\%$ more pallets, and only $12\%$ are using fewer pallets.
These data suggest a positive trend in pallets usage.   

\begin{figure}[t!]
\centering
\subfigure{
\includegraphics[width=0.45\columnwidth]{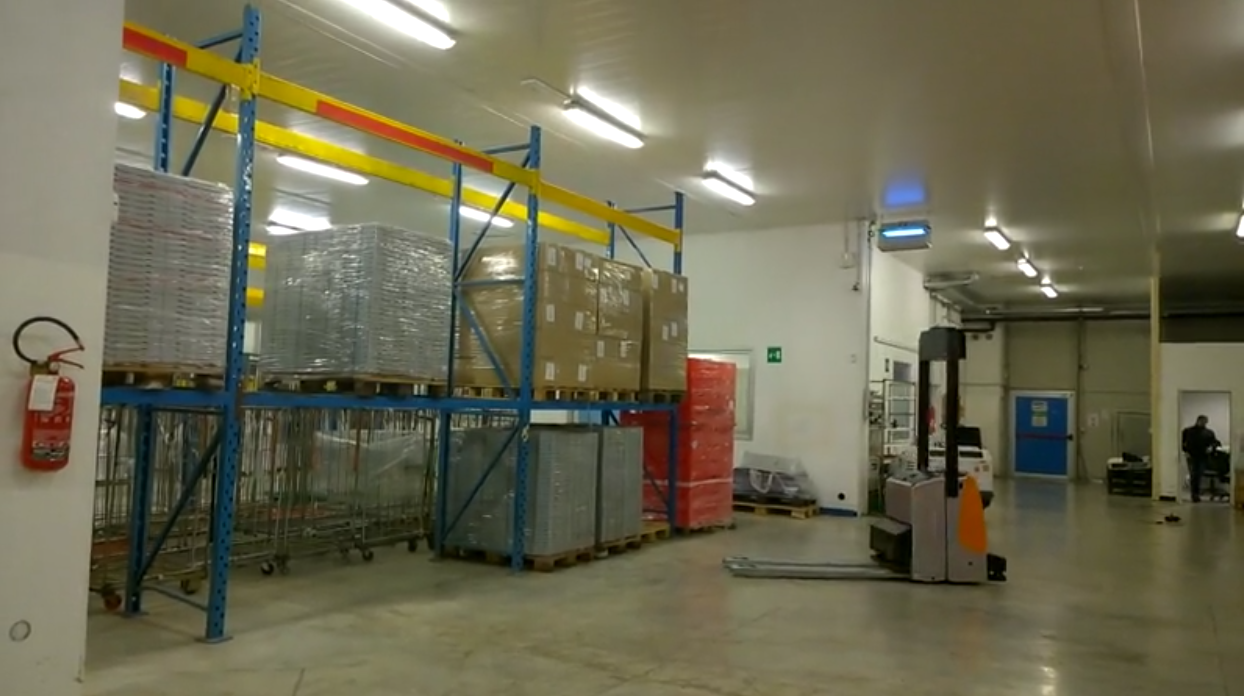}
\label{fig:tortona1}}
\subfigure{
\includegraphics[width=0.45\columnwidth]{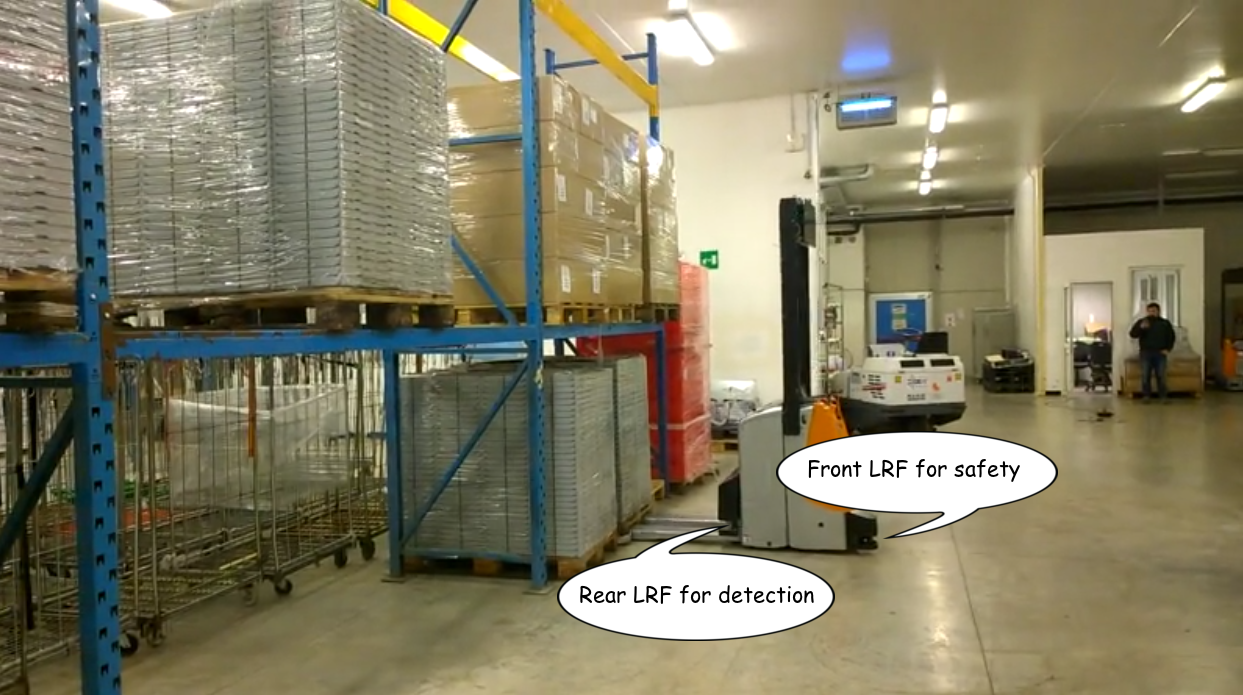}
\label{fig:tortona2}}
\subfigure{
\includegraphics[width=0.45\columnwidth]{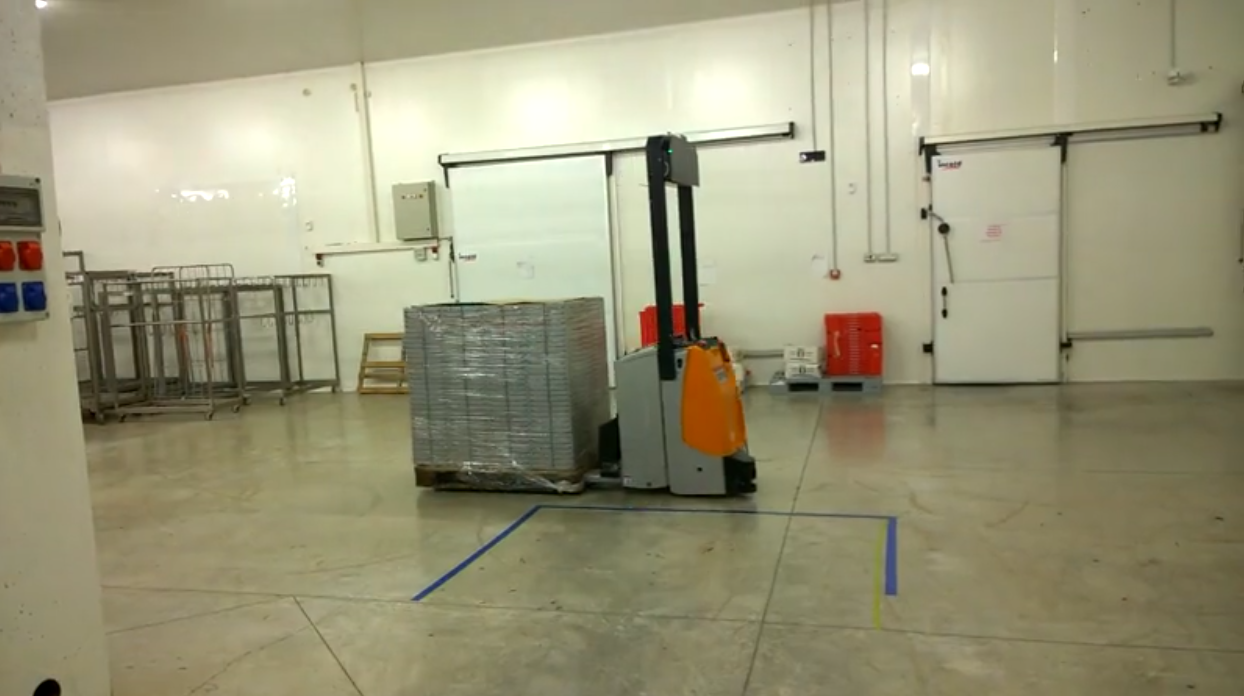}
\label{fig:tortona3}}
\subfigure{
\includegraphics[width=0.45\columnwidth]{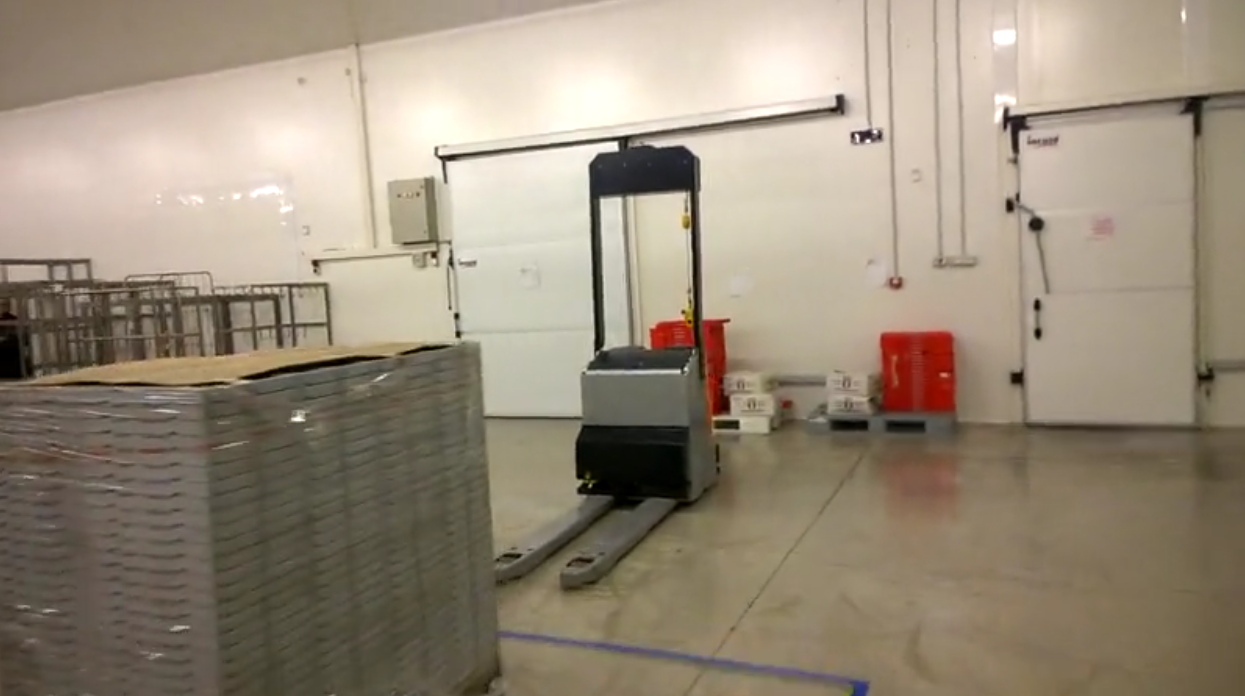}
\label{fig:tortona4}}
\caption{The target experimental scenario: an autonomous forklift designed to approach, fork, transport and place pallets in a warehouse environment in Tortona, Italy.}
\label{fig:transpallet}
\end{figure}

Given all considerations above, the need arises to
(i) provide standard AGVs with the capability of detecting, localising and tracking standard pallets
(ii) when the location of pallets cannot be assumed to be precisely known in advance, and
(iii) in environments where human co-workers operate and other objects are present.
So far, pallet detection, localisation and tracking have received much attention both in scientific literature and in industry-oriented research.
A huge number of studies have been presented, which discuss model-based approaches either adopting computer vision or using 2D laser rangefinder data, and the most relevant ones for this work are discussed in Section \ref{sec:related_work}.
When compared to approaches based on computer vision, 2D rangefinders have the advantages of generating reliable data with a well-characterised statistical sensor noise \cite{pfister2003weighted, Mastrogiovannietal2013a}, being more accurate for long distances, and not being influenced by light conditions. 
However, since laser rangefinders can provide only contour information, they are often coupled with cameras when unique pallet identification is needed \cite{schulenburg2003self}.
As a consequence, the objective of the work described in this paper is two-fold:
\begin{itemize}
\item developing an architecture for commercially available AGVs, in particular forklifts, which has the capability of detecting, localising and tracking (possibly multiple) pallets, using 2D laser rangefinder information; as a target scenario, we refer to the automation of a warehouse located in Tortona, Italy, where a purposely modified commercial forklift has been put in operation, as shown in Figure \ref{fig:transpallet};
\item providing an open, freely available, dataset\footnote{Web: \url{https://github.com/EMAROLab/PDT}.} to the community for further research activities, comprising a collection of $340$ labelled 2D scans related to pallets located in real-world environments \cite{mohamed20192d}. 
\end{itemize}

The major contribution of the paper is an architecture made up of two components:
(i) a pallet detector module employing a Faster Region-based Convolutional Neural Network (Faster R-CNN) detector \cite{girshick2015fast, Ren2017} coupled with a CNN-based classifier for classification purposes operating on a bitmap-like representation of 2D range scans;
(ii) a Kalman filter based module for localising and tracking the detected pallets, as well as increasing the confidence associated with their detection on-line.
In particular, the proposed architecture:
\begin{itemize}
\item to the best of our knowledge, is the first framework for pallet detection, localisation and tracking using machine learning approaches based on 2D range data exclusively; 
\item is designed to detect, localise and track multiple pallets at the same time; 
\item exhibits independence from a possible \textit{a priori} knowledge about a pallet's location;
\item \textcolor{red}{is a data-driven rather than model-based method, in addition to pallets it can be adapted to detect other objects with comparable size and with a similarly fixed geometry;}
\item does not require any modifications to existing standard pallets, as done in other well-known approaches in the literature, for example in \cite{lecking2006variable};
\item does not require information about the forklift's pose either in absolute terms or relative to the target pallet;
\item to the best of our knowledge, this is the first attempt to perform object detection, classification and tracking using a 2D laser rangefinder in conjunction with machine learning methods, instead of the more common model-based approaches. Due to the limited and sparse nature of the data provided by this sensor, this poses different challanges compared to apporaches based on 3D LiDAR, cameras, or both \cite{feng2018towards, zhou2017voxelnet, asvadi2017depthcn, redmon2016you, liang2018deep}.
\end{itemize}

The paper is organised as follows.
Section \ref{sec:background} discusses related work and introduces the reference scenario.
Section \ref{sec:methods} describes the methods to pallet detection, localisation and tracking employed in the proposed architecture.
The overall data flow pipeline as well as the pallet tracking algorithm are described in Section \ref{sec:system_architecture}.
Implementation details and the experimental evaluation are discussed in Section \ref{sec:experimental_results}.
Conclusions follow.


\section{Background}
\label{sec:background}

\subsection{{Related Work}}
\label{sec:related_work}

The problem of designing an autonomous forklift able to fork, transport and place pallets is not new, likewise the problem of pallet detection, localisation and tracking.
Given the geometric shape of a pallet's structure, a number of \textit{model-based} solutions have been proposed in the literature, which make use of either vision or 2D range information, or both. \textcolor{red}{Surprisingly few of them are based on \emph{learning} to recognize the shape of the target object, as opposed to recognizing a known geometrical pattern.}

\textit{{Vision-based systems}}.
A number of vision-based approaches making use of different features extracted from images to detect and track pallets have been presented, and examples include the work described in \cite{byun2008real, chen2012pallet, oh2014development, syu2016computer, holz2016fast, varga2016robust}. 

One of the first approaches to pallet detection and pose estimation has been discussed in \cite{garibotto1997service}.
Soon afterwards, an image segmentation method based on pallet's colour and geometric characteristics has been presented in \cite{pages2001computer}.
However, these approaches require very stable illumination conditions and a very precise camera calibration, which is quite a strong assumption in real-world settings.
The method proposed in \cite{nygards2000docking} attempts to estimate a pallet's pose using a structured light method, which is based on a combination of range and video information.
The main problem associated with such an approach is that its accuracy quickly decreases with distance.
Being able to detect a pallet when it is still distant is a nice-to-have feature in all those cases where pallets are located in a certain load/unload area without a specific arrangement. 
The estimate of a pallet's pose has been attempted also using artificial visual features in the form of \textit{markers} placed on the pallets to detect \cite{seelinger2005automatic, aref2014macro}.
While such approaches do not rely on well-defined illumination conditions, nor they assume a precise camera calibration, it is often difficult to place fiducial markers in real-world environments, because such a process increases setup times to a great extent.

A model-based algorithm using visual information without any fiducial markers or specific illumination conditions has been presented in \cite{garibott1996robolift}.
This algorithm exploits the identification of a pallet's central cavities to identify two pallet slots and estimate their geometric centre in calibrated images.
However, such a system requires an accurate \textit{a priori} knowledge of a pallet's pose, which (as described above) is not realistic in real-world settings.
A retrofitted autonomous forklift with the capability of stacking racks and fork pallets placed within a certain area with uncertainty was presented in \cite{kim2001model}.
The docking method for pallet forking is based on the detection of specific reference lines for concurrent camera calibration and pallet identification, and it allows for the stacking of well-illuminated racks and the localisation of pallets in front of the vehicle.
Unfortunately, such a solution proves to be limited to the stacking task only. 
The approach described in \cite{Cucchiara2000FocusBF} is based on a more complex visual processing pipeline, which employs a number of hierarchical visual features like regions, lines and corners using both raw data and template-based detection.
In \cite{wang2016autonomous}, the authors present an autonomous pallet handling method based on a line-structured light sensor, where the design of such a sensor is based on an embedded image processing board containing an FPGA and a DSP.
This approach can identify and localise pallets using their geometrical structure based on a model-matching algorithm, and uses a position-based visual servoing method to drive the vehicle while it approaches the pallet to fork.
Unfortunately, it also requires the development of custom hardware. 

An approach for the automated pallet detection combining stereo reconstruction and object detection from monocular images has been presented in \cite{varga2014vision}.
Improvements and extensions for a stereo camera system responsible for autonomous load handling were presented, by the same authors, in \cite{varga2015improved}.
However, the use of stereovision and structure-from-motion algorithms can hardly fit with real-time requirements typically needed when autonomous vehicles are present. 
The work described in \cite{cui2010robust} introduces a method to identify pallets using color segmentation in real time.
However, such a method is prone to the presence of false positives, unless assumptions about pallets colour are posed.
A comparison between two common 3D vision technologies, namely the photonic mixer device (PMD) and typical stereo camera systems, was presented in \cite{beder2007comparison}.
The authors conclude that the PMD system is characterised by a greater accuracy than a typical stereo camera system.
On the basis of such an insight, a solution for pallet loading and de-palletising detection employing a PMD camera has been introduced in \cite{weichert2013automated}.
Again, such approach requires the introduction of \textit{ad hoc} hardware, at the expense of cost and maintenance.

Overall, vision-based systems are characterised by a number of drawbacks, which make their use still limited to specific conditions, including: (i) the need for fiducial markers or similar mechanism to reduce false positives; (ii) the need for stable environmental conditions; (iii) computational load of the associated computer vision algorithms; (iv) the need for custom hardware solutions to enable real-time operations.

\textit{{Rangefinder-based systems}}.
Traditionally, 2D laser rangefinders have been extensively employed for robot localisation and mapping, and such techniques have been also successfully applied to environments characterised by a high degree of human presence \cite{Mastrogiovannietal2007, Mastrogiovannietal2008, Capezioetal2011}.
In the past few years, a number of model-based approaches have been presented and discussed in the literature, which constitute effective methods to detect, localise and track pallets in range data.
In contrast to vision-based algorithms, such approaches do not suffer from image distortions (related to camera calibration), varying illumination issues or object scaling problems, which can result in false detections or mis-detections of significant features, and are characterised by lower computational requirements.
The early work by Hebert \textit{et al}. \cite{hebert1986outdoor} describes techniques for scene segmentation, object detection, and object recognition with an outdoor robot using range data.
In \cite{hoffman1987segmentation}, the authors present a method for detecting and classifying objects using range information.
A model-based technique that leverages prior knowledge of an object's surface geometry to jointly classify and estimate the surface structure was proposed in \cite{newman1993model}.
However, such models are characterised by bold assumptions related to perfect data association and absence of noise.    

Starting from these initial results, range data have been applied to the pallet detection problem.
Data acquired from a laser rangefinder are used in \cite{baglivo2008object} to detect and localise pallets, but the approach cannot deal with ambiguous matches, i.e., it requires perfect data association.
The solution discussed in \cite{walter2010closed} uses a fast linear programming method for detecting line segments in range data, as applied to pre-filtered points selected by a human using an image provided by a camera mounted on a forklift.
In particular, pallets are identified by the classification of detected line segments belonging to its front, and their position is therefore computed.
However, such a method requires a pre-processing step, and its precision can be hampered in the case of specific pallet poses.
In \cite{lecking2006variable}, the authors present two approaches based on 2D range data:
the former assumes the availability of pallets modified with reflectors to compute their position and orientation, whereas the latter uses only their geometrical characteristics as it may be unfeasible to place reflector marks in all pallets.
In the second case, the Iterative Closest Point (ICP) algorithm is used to match range data with the pallet model.
However, the main drawback of the approach is that ICP needs an initial (although approximate) pallet's location, otherwise the iterative computation can become very time consuming and leading to inaccurate results. 
As discussed above, this may be unrealistic in real-world situations.
The work presented in \cite{he2010feature} discusses a feature-to-feature matching for pallets, which first detects line segments, and then matches them with the pallet's geometric model.
However, such an approach can lead to a number of false positives and to ambiguous pose estimations.
Other methods for 2D data segmentation, feature detection, fitting, and matching have been presented in \cite{premebida2005segmentation, bostelman2006visualization}, but all these approaches are characterised by the same drawbacks.
An integrated laser and camera sensory system, for solving the problem of simultaneously identifying and localising pallets whose location is characterised by a great uncertainty, has been presented in \cite{baglivo2011autonomous}.
However, such an approach suffers from a number of drawbacks typically associated with vision processing.

In summary, rangefinder-based systems avoid certain limitations associated with vision-based approaches, but are nonetheless limited as far as detection capabilities are concerned, such as: (i) the need for a model-based approach grounded on pallet geometry; (ii) the necessity of computing features enabling model matching processes; (iii) the ease at which detection estimates can diverge.

\textit{{Discussion}}.
At a first glance, model-based approaches seem appealing because pallets are characterised by a well-defined shape and geometrical features. 
However, in order to enable a reliable and robust detection, experience suggests that many assumptions are to be made.
In fact, all the approaches in the literature, in a way or another, are characterised by recurring limitations: 
while vision-based methods are highly dependent on light conditions (or assume them to be stable), camera calibration issues, and pallet-to-camera distance, or assume to retrofit the environment with the adoption of fiducial markers on each pallet, 
range-based methods are based on models grounded on pallet's geometry, and require the stable detection of certain characterising features. 
From our analysis, it appears that the use of machine learning techniques for the problem of pallet detection and localisation, coupled with solutions for pallet's pose tracking, and when only range information is used, has not been explored in the literature.
Such an approach is expected to avoid the drawbacks associated with vision-based approaches, does not assume any \textit{a priori} model for pallets, does not compute high level features, and \textcolor{red}{can be potentially ported to other similar applications via retraining of the object recognition module.
In addition, the method includes} a sequential classification procedure to reduce the occurrence of false positives.
Furthermore, it can be seen generally that the most used sensors for object detection, classification and tracking based on machine learning techniques are 3D LiDARs \cite{feng2018towards, zhou2017voxelnet,asvadi2017depthcn}, cameras\footnote{Web: \url{http://cs-chan.com/source/FADL/Online_Paper_Summary_Table.pdf}} or a combination of LiDARs and cameras \cite{liang2018deep, matti2017combining}, while 2D laser rangefinders are usually avoided for this task despite their convenience, as they provide only partial contour information. This fact poses the challenge of how to make use of sparse data with limited information content, while still achiving a system with robust tracking capabilities and a small number of false positive detections.

\subsection{The Reference Scenario}
\label{sec:reference_scenario}

The scenario we target in this paper includes a purposely modified model EXU low lift pallet truck manufactured by STILL GmbH, which has been put in operation in a warehouse environment in Tortona, Italy.
The forklift, depicted in Figure \ref{fig:transpallet}, can lift up to $2.200$ $Kg$ at a $760$ $cm$ height.
It has been equipped with two safety laser rangefinders \textcolor{red}{operating at $16$ $Hz$} for obstacle detection \textcolor{red}{and safety purposes}, placed as to cover a full $360$ $deg$ scan around the truck, and one of them, namely the rear sensor as shown in Figure \ref{fig:transpallet}, can be used to provide the data required by our architecture.
Furthermore, it has been extended with a localisation system performing tri-lateration using a number of intelligent devices distributed in the environment \cite{Mastrogiovannietal2009a, Mastrogiovannietal2010, Capezioetal2011, Mastrogiovannietal2013a}.
Being able to localise and avoid obstacles, the forklift can freely move in the warehouse.
The map of the environment is assumed to be known (either available \textit{a priori} or built off-line), and a number of relevant locations, such as forking and placing areas, are identified as semantic tags in the map.
It is noteworthy that forking and placing areas are roughly $3 \times 3$ $m^2$ regions where a pallet can be located anywhere inside it. 
Therefore, it is not possible to assume in advance a specific location or pose for the pallet, as it is typically done by other approaches in the literature described above, but it can be fairly assumed that it lies within the area. 
\textcolor{red}{Once the pallet truck gets close to the placing and forking areas, it slows down to a suitable speed for safe pallet loading and unloading, namely $0.2$ $m/s$.}

Missions are defined using a knowledge representation and planning framework previously developed for mobile robots \cite{Mastrogiovannietal2004}.
The framework is able to express a high-level goal in an ontology-based representation, and to determine a corresponding set of planning problems, whose solutions (i.e., plans) are guaranteed to achieve the goal, if feasible.
Missions are typically configured as sequences of forklift motions to a given goal location, approaching the pallet to fork, forking, delivering the pallet to the placing area.
Once the forklift moves towards the forking area, it needs to detect, localise and then track the pallet (which is still) to compensate its own motion. 
As per functional requirements of the use case we consider, only one pallet is present in the forking area, and therefore the specification is related to the detection of one pallet only.
In the paper, we also consider the case in which two pallets may be present in the forking area at the same time, in order to better discuss the capabilities of our approach.
In principle, given the forklift's localisation system, tracking would not be necessary, as pallets do not move and robot motion information could be used for pallet localisation. 
However, in our case we decided to determine whether it is possible to perform pallet tracking without relying on the robot's localisation capabilities, therefore taking inspiration from the literature on the use of minimal information for localisation and navigation in mobile robotics \cite{Mastrogiovannietal2009b}.

Obviously enough, the problem has been already explored in the literature.
In particular, the work described in \cite{aref2013position, aref2014macro} integrates visual information with robot's odometry to implement a smooth and non-stop transition from autonomous navigation to visual servoing.
However, in order to perform such a transition, it is necessary to understand when visual servoing can be activated to avoid scattering motions.
When using commercially available rangefinders, the typical maximum pallet detection range from a robot is approximately $4$ $m$ \cite{aref2016multistage}.
A few systems include multiple-view rangefinders, and therefore are capable of attaining longer detection distances in the forklift's workspace, i.e., up to $6$ $m$ \cite{walter2010closed, walter2015situationally}.
As we better discuss below, being able to detect a pallet from longer distances is a nice-to-have feature when sequential classification processes are employed.

In our scenario, the forklift does not employ any specific strategy to approach pallets.
When moving towards the picking area, pallet detection, localisation and tracking are activated, \textcolor{red}{and the forklift slows down to a speed suitable for pallet loading and unloading.
At this point, 2D range scans from the sensor are processed by the proposed algorithm in order to detect pallets. It is noteworthy that such data is acquired at all times independently from our algorithm, manily for safety reasons, and that a pallet detecion and tracking solution may consume such data at a lower frame rate than the one provided by the sensor, if the algorithm is robust enough and accordingly with the vehicle operating speed.}

\textcolor{red}{In details, we focus on standard EUR/EPAL pallets, whose dimensions are $120$ $cm$ $\times$ $80$ $cm$ $\times$ $14.4$ $cm$.}
Multiple range scans are subsequently used to localise and track pallets, and to remove false positive detections.
When a sufficient confidence level on a tracked pallet is reached, then the pallet is considered successfully detected. 
Such a sequential classification process can benefit from the fact that, when the forklift is approaching the forking area, it can already ascertain whether a pallet is present and where it is located.
The whole process is described in details in Section \ref{sec:system_architecture}.

\section{Methods}
\label{sec:methods}

\subsection{Convolutional Neural Networks}

{\color{red}
Convolutional Neural Networks (CNNs) are specifically suited for image processing applications \cite{Lecun2015}. They are specialised multi-layer perceptrons that include a \emph{hard inductive bias} in the form of a strongly constrained structure that is especially suited to signal processing. To this purpose, the main component of CNNs is convolutional layers. These layers feature a local connectivity patterns, forcing the network to operate on limited-size receptive field. Since the weight values are repeated over each of these receptive fields, the result is that a trained convolutional layer effectively implements the convolution of its input signal with a learned filter. Nonlinear layers are then used to decimate the output of multiple filter banks and to provide the capacity of learning more general non-linear mappings.

In addition to being tailored to signal-processing applications, this constrained structure is necessary to avoid an unmanageable number of parameters to train even for small sized images.} Since each filter is defined only by a small number of parameters compared to a fully connected layer, the number of parameters to be trained in the network is greatly reduced.

A number of hyper-paramenters are should be set when designing a convolutional layer, such as:
(i) its \emph{depth}, i.e., the number of filters to be used;
(ii) the \emph{receptive field}, i.e., the size of each filter;
(iii) the \emph{stride}, being how much to slide the filters; and finally
(iv) the \emph{zero-padding}, i.e., how many zeros to pad a layer's input with, so to control the output's size. 

A CNN is usually composed of more than a single convolutional layer, and often includes other kinds of layers.
A typical structure involves:
\begin{itemize}
\item an \emph{input layer}, having the same dimensions as the input data (e.g., a colour image having dimensions $250 \times 250 \times 3$ pixels by colour channels);
\item the \emph{convolutional layer}, as described in the previous paragraph, usually increasing the depth of the volume by computing multiple filters;
\item a \emph{Rectifier Linear Unit (ReLU) layer}, which applies an element-wise nonlinear activation function and keeps dimensions unchanged;
\item a \emph{pooling layer}, to perform downsampling on spatial dimension (i.e., width and height);
\item a final \emph{fully connected layer}, to classify the input data based on the previously computed features, i.e., a vector with length equal to the number of possible classes.
\end{itemize}
\textcolor{red}{Not all of these layer types have learnable parameters, as ReLU and pooling layers do not, but all of them except ReLU are characterised by some hyper-parameters to be defined.}

Based on the advancements in CNNs, Region-based Convolutional Neural Networks (R-CNNs) have been proposed to perform object detection tasks \cite{girshick2014rich, girshick2015fast, Ren2017}, i.e., the task of associating a number of \emph{bounding boxes} with an image, each one possibly corresponding to (i.e., enclosing in image space) an object of interest.
The R-CNN family includes R-CNNs \cite{girshick2014rich}, Fast R-CNNs \cite{girshick2015fast} and Faster R-CNNs \cite{Ren2017}.
In general, these region-based approaches are organised as a 2-step process:
they generate a set of bounding box proposals, and 
submit those regions of interest to a classifier to determine whether any of them is an object (i.e., their \emph{objectness}) and which object they correspond to.
In brief, R-CNNs and Fast R-CNNs rely on external region proposals generated by Selective Search \cite{uijlings2013selective}, and present a rather complex training pipeline.
On the contrary, Faster R-CNNs add a fully convolutional layer on top of the features maps generated by the last convolutional layer, called \emph{Region Proposal Network} (RPN).
The RPN works by passing a $n \times n$ sliding window over a set of convolutional feature maps on the last convolutional layer, so as to propose bounding box candidates of predefined scales and aspect ratios.
RPN defines a number of region boxes in the image space (called \textit{anchors}) and ranks them on the basis of their likelihood of containing objects, in our case pallets.
As it is customary in Faster R-CNNs, for each sliding window on the convolutional feature map $9$ anchors are generated with $3$ different sizes and 3 different aspect ratios in all possible combinations \cite{girshick2014rich, girshick2015fast, Ren2017}, all of them compatible with standard EUR pallets.
Furthermore, for each anchor a value $o$, is computed, which refers to the overlap ratio between the areas of anchors and of \textit{ground truth} bounding boxes:
\begin{equation}
o =
\left\{
\begin{array}{lcl}
1 \;\;\rightarrow \text{Decision: Object}		& \textit{if} \;\; \textit{IoU}> 0.7,\\[2ex]
0 \;\;\rightarrow \text{Decision: Not an object} 	& \textit{if} \;\; \textit{IoU}< 0.3,
\end{array}
\right.
\label{eq:ovalue}
\end{equation}
where \textit{IoU} (which stands for \textit{intersection over union}) can be defined as:
\begin{equation}
IoU = \displaystyle \frac{\displaystyle Area(anchor) \cap Area(ground\;truth\;bounding\;box)}{\displaystyle Area(anchor) \cup Area(ground\;truth\;bounding\;box)}.
\end{equation}
In \eqref{eq:ovalue}, the thresholds over $IoU$ can be tuned experimentally.
 Eventually, these features are then fed to a network with two main tasks, namely regression and classification.
The regression output determines the predicted bounding boxes, each with a form of $[x_{min},y_{min}, x_{len}, y_{len}]$, while the output of the classification network is the value $o$ indicating whether each predicted bounding box contains an object, according to \eqref{eq:ovalue}.

This implies that Faster R-CNNs achieve efficient and fully end-to-end training, as a single CNN is used for region proposal and classification.
Hence, Faster R-CNNs address the limitations of other architectures and achieve greatly improved performance, being much faster than regular R-CNNs. 

\subsection{Sequential Classification}

We have approached the problem of detecting a pallet across multiple 2D scans as a sequential decision problem.
Sequential decision \cite{Wald1945, Ghosh1970} is an approach to classify multiple observations on the basis of assigned confidence intervals.
When the joint probability of the observed sequence is \emph{sufficiently} high or low to escape from an uncertainty range, the decision is taken.

The simplest sequential analysis method applies a Bayesian analysis to compare the joint class-conditional probabilities of the observations so far, by evaluating their ratio.
If, at observation number $i$, the posterior probabilities of \emph{Class 1} (e.g., a pallet is present) and \emph{Class 0} (no pallet is present) given the observations $x_1$, $\ldots$, $x_i$ are $p_{1}(i)$ and $p_{0}(i)$ respectively, and if $A$, $B$ ($A > B$) are two thresholds related to the balance between errors related to false positives and to false negatives, then the decision criterion is:
\begin{equation}
\left\{
\begin{array}{lcl}
\dfrac{p_{1}(i)}{p_{0}(i)}\ge A 	& \rightarrow \qquad& \text{Decision: Class 1,}\\[2ex]
\dfrac{p_{1}(i)}{p_{0}(i)}\le B 	& \rightarrow & \text{Decision: Class 0,}\\[2ex]
B<\dfrac{p_{1}(i)}{p_{0}(i)}<A 	& \rightarrow & \text{Decision: continue observing the sequence.}
\end{array}
\right.
\label{eq:sequential}
\end{equation}

In the original formulation, the probabilities are assumed to be known and observations to be mutually independent, so if at step $i$ the class-conditional probabilities of the current observation $x_i$ are $f_1(x_i)=\Pr(\text{Class}=1\;\vert\;x_i)$ and $f_0(x_i)=\Pr(\text{Class}=0\;\vert\;x_i)$, respectively, we can write the basic \emph{sequential probability ratio} as:
\begin{equation}
\label{EQ:sprt}
\frac{p_1(i)}{p_0(i)}
\;=\; \frac{f_1(x_1) f_1(x_2) \cdots f_1(x_i)}{f_0(x_1) f_0(x_2) \cdots f_0(x_i)}
\;=\; \prod_{j=1}^i\dfrac{f_{1}(x_j)}{f_{0}(x_j)}.
\end{equation}

In the present case, this method has been applied with scores generated by a soft classifier rather than true probabilities, as described in the previous Section.
More importantly, the assumption of independence is not realistic when considering subsequent 2D scans \textcolor{red}{ at a high refresh rate, as the ones obtained from a laser rangefinder}.
Methods taking into account dependencies, for instance under a Markov assumption \cite{Novikov2001}, are available.
These have a higher computational time, possibly incompatible with real-time operation.
Moreover, they have a higher number of parameters, since they explicitly model the expected dynamics (for instance as a Markov chain), so they have higher sample complexity and are more prone to overfitting.
It should be noted that the independence assumption in this context is safe, although maybe suboptimal, since it gives a worst-case estimate.
As shown in Section \ref{sec:experimental_results}, this worst-case approach proved to yield good results, so considering its computational and learning complexity advantages this was the preferred approach.


\begin{figure*}
\renewcommand{\figurename}{Fig.}
\centering
\includegraphics[width=0.9\textwidth]{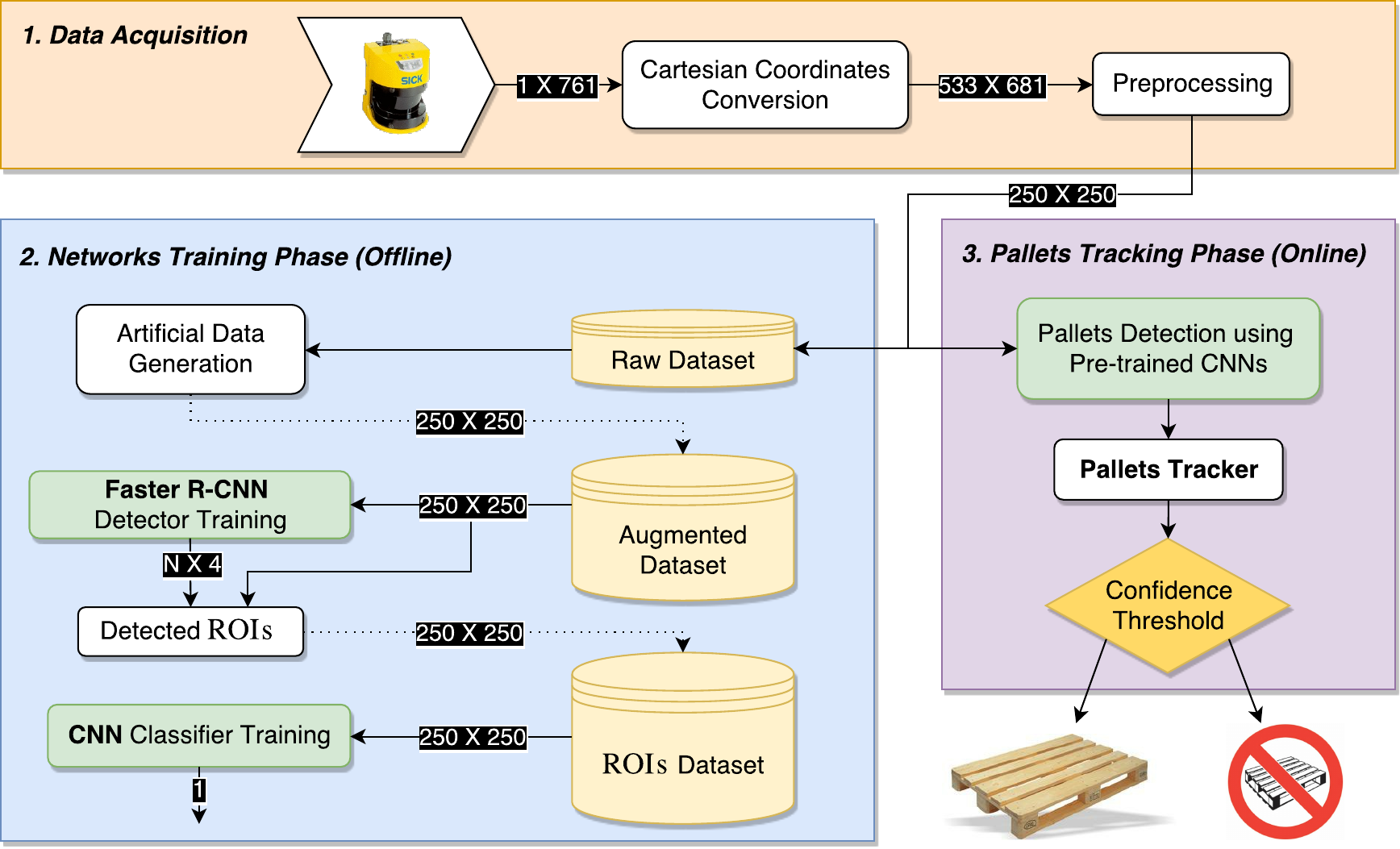}
\caption[An overview of the proposed system for pallet detection, localisation, and tracking.]
{An overview of the proposed system for pallet detection, localisation and tracking. Black labels on arrows represent the size of exchanged data.
}
\label{fig:pipeline}
\end{figure*}
\section{System's Architecture}
\label{sec:system_architecture}

In this Section, we discuss the structure of the proposed architecture, which is depicted in Figure \ref{fig:pipeline}. 
{\color{red}
It consists of three parts.
First (\emph{Phase 1}), raw range data are acquired, and each scan is converted into a 2D bitmap-like image, so that it is in the most appropriate format for a CNN.
Then  (\emph{Phase 2}), a dataset of real-world 2D scans, each one converted into a bitmap, is collected and \emph{offline} training of a Faster R-CNN is performed. The network will detect regions of interests (ROIs), and eventually classify them as pallet candidates using a separate CNN.
Once training is complete  (\emph{Phase 3}), the pallet detection module trained in the previous step is coupled with a Kalman filter-based tracker, which is used \emph{online} to match potential pallet detections over time. The novelty of this step is that, instead of immediately accepting a potential pallet, the decision can be deferred until sufficient confidence in the candidate is achieved, reducing the chance to pursuit a false positive and stabilising true positives in case of a temporary occlusion or sensor noise. On the other hand, if the candidate's confidence falls below a certain threshold or the it disappears for a several frames, the candidate pallet is just discarded.
}

\subsection{Data Acquisition}

A single laser rangefinder scan $s_i$ taken at the time instant $i$ can be represented using a set of polar coordinates:

\begin{equation}
s_i = \lbrace(r_1, \phi_1), \ldots, (r_j, \phi_j), \ldots, (r_M, \phi_M)\rbrace,
\end{equation}	
being $M$ the number of single range points, i.e., related to the angular sensor's resolution.
Hence, $r_j$ is the measured distance of an object with respect to the rangefinder location in the direction given by the angle $\phi_j$.
For a single point in range data, we can obtain a binary image of the operating area's floor plan, converting the acquired data to Cartesian coordinates with the following formula:
\begin{equation}
\left\{
\begin{array}{l}
x_j = r_j\cos(\phi_j),\\
y_j = r_j\sin(\phi_j).
\end{array}
\right.
\;
\end{equation}
This second representation is preferred for object detection and tracking as it allows for recovering correlations among neighbouring pixels in 2D images, which can be exploited by the CNN layers.
In particular, we first convert data from a laser rangefinder, which has been limited to $6$ $m$ maximum depth, into $533 \times 681$ pixel binary images.
Such images are then resized to $250 \times 250$ pixels, leading each pixel to cover an area of $4.5$ $cm^2$.
Such discretisation has been deemed sufficient to take into account motion noise during pallet forking actions.

\begin{figure}
\renewcommand{\figurename}{Fig.}
\begin{center}
\includegraphics[width=0.9\columnwidth]{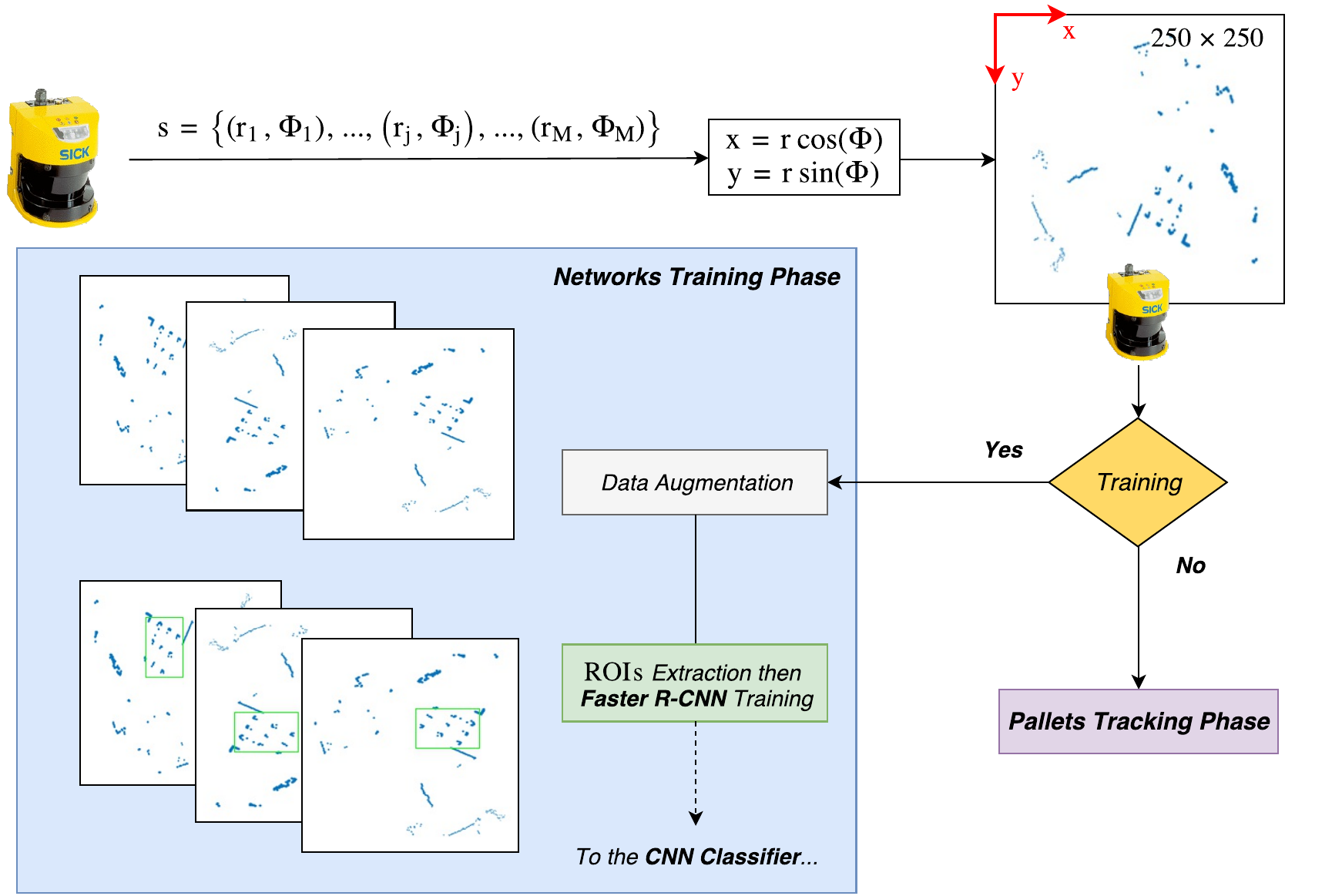}
\caption[The main steps of the data preparation phase.]
{The main steps of the data acquisition (preparation) phase.}
\label{fig:prephase}
\end{center}
\end{figure}
When such operation is done, images are ready to be used for online detection, localisation and tracking.
However, two additional steps are required to prepare the necessary datasets for training purposes, namely the use of artificially generated images and the definition of specific regions of interest (ROIs).
ROIs are 2D bounding boxes of the objects (e.g., pallets) in the dataset images and, as described above, are defined by their top-left and bottom-right corner points, i.e., $(x_{min} , y_{min})$ and $(x_{max}, y_{max})$, which uniquely identify a region's position and size.
First, the available dataset of real-world data is augmented with artificially generated images.
The new images can be obtained by translations and rotations of the original ones, in order for the network better generalise with respect to pallet locations and poses.
The generation of artificial data with the aim of reducing overfitting in image-based training has already been used \cite{teng2010real, brust2015convolutional}.
Such a technique also has the advantage of reducing the time and efforts devoted to:
(i) collecting a large amount of real-world data, and
(ii) labelling such data with \emph{ground truth} so that they can be used for training, as it is possible to infer the new labels whilst the corresponding data is generated. 
This first dataset is used to train the Faster R-CNN detector.
Once this first network has been trained, it can be used to extract the ROIs associated with the objects in the dataset.
The ROIs dataset, as it is described in Section \ref{sec:experimental_results}, is then used to train the CNN-based classifier, that detects which ROIs may correspond to a pallet.
A summary of these steps can be found in Figure \ref{fig:prephase}.

\subsection{Neural Networks for Pallet Detection}

In order to track pallets in 2D images obtained via range data, we need first an approach to reliably detect them in each single image.
As anticipated, we designed such module using neural networks.
This Section is focused on the general architecture of such neural networks.
Details on the training process for our specific experiments, such as size and composition of datasets, are given in Section \ref{sec:training}.

The pallet detection process is made-up of two steps: 
a state-of-the-art Faster R-CNN detector which detects the ROIs in each image, and a CNN-based classifier taking as input the previous step detections and discriminating which of them could be a possible pallet candidate.
In the first step, we use a Faster R-CNN for two reasons:
(i) it allows for detecting possibly multiple pallets while being robust to false positives, and in this sense the Faster R-CNN provides us with a number of ROIs that can undergo further inspection in the second step;
(ii) we want to estimate the position of each detected pallet, and the centroid of an associated ROI can be used to that purpose.
However, the Faster R-CNN is not sufficient for a reliable identification of each ROI, and this is why a CNN-based classifier is necessary.
It is noteworthy that in so far as the 2D laser rangefinder is a robust and reliable sensor, the amount of data it provides is limited to partial objects' contours on a plane.
A CNN-based classifier is able to detect pallets in a ROI with a small number of false positives despite the limited amount of cues.
As anticipated above, the two networks are completely independent, and are trained with different training sets.

The Faster R-CNN detector is composed of several layers, divided in three main stages: the input layer, an intermediate convolutional stage, and a final fully connected stage.
The input layer consists of the input image corresponding to the 2D scan, downscaled to a $32 \times 32$ pixel grey-scale or RGB images to improve general performance.
The central convolutional stage is made up of two convolutional layers, interleaved by two ReLU layers, and followed by a final max-pooling layer.
Each convolutional layer applies $40$ filters, with a size of $3$ and a stride and a padding of $1$, whereas the max-pooling layer employs pooling regions of size $3$ and a stride of $1$, which produces output images of size $30 \times 30$.
The final stage is composed of two fully connected layers, followed respectively by one ReLU layer and a softmax classification layer.
The first fully connected layer is composed of $64$ neurons, which is followed by the ReLU layer. 
The output size of this layer is an array with a length of $64$, which represents the most significant features in the image. 
Such features are then used by the last fully connected layer combined with the softmax classification layer to determine whether a ROI proposed by the RPN belongs to one of the object classes (i.e., pallets) or to the background, using sequential classification.
The overall output is a list of candidate ROIs, defined by two corner points as described in the previous Section.

The CNN-based classifier that follows is trained to classify the most promising ROIs detected by the Faster R-CNN as pallets.
The classifier is trained using a dataset obtained based on the ROIs bounding boxes and the original images, as detailed in Section \ref{sec:training}.
Compared to the first network, the CNN-based classifier is simpler in its structure.
The input layer gets as input filtered full-size images so that they contain only the ROIs, and therefore it has a $250 \times 250$ size.
The input layer is then followed by a convolutional layer, a ReLU layer, and a max-pooling layer.
The convolutional layer has depth of $25$, filter size of $20$, and stride and padding set to $1$, whereas the max-pooling layer applies pooling regions of size $5$ and stride equal to $2$.
Finally, a fully connected layer and a softmax classification layer are employed to compute the salient features and classify the image accordingly. Eventually, the performance of the proposed networks is evaluated by empirical validation with $k$-fold cross-validation, as it provides a reliable assessment of the network accuracy \cite{arlot2010survey} without excessive burdening on computation time.

\subsection{Online Pallet Tracking}
\label{sec:pallettracking}

The two networks described in the previous Section are trained to detect ROIs as bounding boxes and determine which of them correspond to a pallet.
It is tempting to assume that the pallet detection problem is solved and consider tracking just as a step necessary to approach pallets.
We argue that deferring the decision about pallet detection until a sufficient confidence level is reached is a wiser approach, and that tracking should play an important role in the detection process.
The aim is to avoid all those situations where a single spurious sensor reading can mislead the system towards a false positive or immediately give up on a promising candidate pallet.
Consequently, we do not immediately accept a certain ROI as a true positive pallet classification, but rather like a \emph{candidate} that must be validated.
This can be achieved adopting a sequential classification approach, i.e., by tracking all candidate pallet detections but taking a final decision only later, when different scans have been acquired and the system has gained sufficient confidence. 

As it is typical when performing object tracking, the overall tracking process involves two steps:
\begin{itemize}
\item detecting candidate pallets in each single frame using the pre-trained networks as perception models, therefore obtaining a set of ROIs;
\item performing data association, i.e., associating ROIs related to the same pallet over different scans, which we refer to as a \textit{track}.
\end{itemize}

In our system, data association for each track is based on perceived track displacements.
Assuming that the robot is moving with a constant velocity, each detected pallet in the current frame is located in a slightly different location with respect to the previous one.
In reality, pallets are still, but robot perceptions create such illusion of displacement because of ego-motion effects.
Data association is achieved by estimating the motion of each candidate pallet over several frames using a linear Kalman filter \cite{cuevas2005kalman, mohamed2017detection}. 
The filter is used to predict the position of the centroid of each track in the current frame based on past positions, and the corresponding bounding box is updated accordingly.
In the current version of the system, rotations are not considered.
Then, the associations between ROIs and tracks are computed and ranked. 
The association is done using the Hungarian algorithm \cite{kuhn1955hungarian}.
The algorithm minimises a cost function computed using the overlap between the bounding box location as predicted by the Kalman filter, and the bounding box detected by the pre-trained networks.
The minimum is achieved when the predicted bounding box is perfectly aligned with the detected bounding box, i.e., the overlap ratio is one.
In any frame, some of the ROIs may be assigned to tracks, while others may remain unassigned. 
At the same time, already available tracks may not be associated with any ROIs in the current scan.
Estimated centroids of assigned tracks are updated using the corresponding ROIs inversely weighted by the corresponding confidence values acting as covariance matrices, while unassigned tracks are not updated.
It is foreseen that each unassigned ROI originates a new track. 
In order not to propagate older tracks, each track is associated with a counter related to the number of consecutive frames where no associations have been made with such a track, and to the recent average confidence values. 
If such a counter exceeds a specified threshold or the average score associated with the ROI's likelihood of being an object is below a certain threshold, the algorithm assumes that the pallet associated with the ROI is no longer in the rangefinder's field of view, or it is a false positive, and therefore it deletes the track. 

The data acquisition process is better described in Algorithm \ref{alg:acquisitionphase}, which employs the two networks described above.
The set $\mathbb{C}$ of candidate pallets (i.e., the corresponding ROIs) is first initialised (line 2).
A scan is acquired (line $3$), and converted to a bitmap-like image (line $4$).
Then, the Faster R-CNN detects all possible ROIs (line $5$).
The neural network is embedded in a function called \textsc{DetermineROIs()}. 
For each ROI $r_i$, if the associated \emph{objectness} score $o_i$ is above a given threshold $\epsilon_{objectness}$, then $r_i$ is passed down to the CNN-based classifier in function \textsc{Score()} to compute the associated confidence score $cs_i$ (line $9$). 
If such a score is higher than a threshold $\epsilon_{candidate}$ (line $7$), then $r_i$ is included in the set $\mathbb{C}$ of candidate pallets.

\begin{algorithm}[th!]
\caption{Acquisition of candidate pallets}
\label{alg:acquisitionphase}
\begin{algorithmic}[1]
\Function{Acquisition()}{}
\State $\mathbb{C} \leftarrow \emptyset$
\State $s_i \leftarrow$ \textsc{AcquireScan()}
\State $i_i \leftarrow$ \textsc{Convert($s_i$)}
\State $\mathbb{R} \leftarrow$ \textsc{DetermineROIs($i_i$)} 
\ForAll{$r_i \in \mathbb{R}$}
	\If{$o_i > \epsilon_{objectness}$}
		\State $cs_i \leftarrow$ \textsc{Score($r_i$)} 
		\If{$cs_i > \epsilon_{candidate}$}
			\State $\mathbb{C} \leftarrow \mathbb{C} \cup r_i$
		\EndIf
	\EndIf
\EndFor
\State \textbf{return} $\mathbb{C}$
\EndFunction
\end{algorithmic}
\end{algorithm}

\begin{algorithm}[th!]
	\caption{Pallets tracking}
	\label{alg:trackingphase}
	\begin{algorithmic}[1]
	\Function{PalletsTracking()}{}
	\State {\textbf{Requires} the velocity $V$ of the robot, a time window $W$}
	\State $\mathbb{T} \leftarrow \emptyset$
	\State $\mathbb{D} \leftarrow \emptyset$
	\State $\mathbb{U} \leftarrow \emptyset$
	\Loop
		\State $\mathbb{C} \leftarrow$ \textsc{Acquisition()} 
        		\For{$t_i \in \mathbb{T}$}
			\State $d_i \leftarrow$ \textsc{RetrieveROIs($t_i$)}
			\State $S_{avg_i} \leftarrow$ \textsc{RetrieveAverageConfidence($t_i$)}
			\State $p_i \leftarrow$ \textsc{RetrievePose($t_i$)}
			\State $p_i \leftarrow$ \textsc{KalmanPrediction($p_i$, $V$)}
		\EndFor
		\For{\textbf{all} $c_j \in \mathbb{C}$}
			\For{\textbf{all} $t_i \in \mathbb{T}$}
				\If{\textsc{Area($c_j \cap t_i$) $<$ $\epsilon_{overlap}$}}
					\State $p_i \leftarrow$ \textsc{KalmanUpdate($p_i$, $c_j$)}
					\State $d_i \leftarrow d_i + 1$
					\State $S_{avg_i} \leftarrow$ \textsc{Update($W$)}
					\State $\mathbb{U} \leftarrow \mathbb{U} \cup t_i$
				\Else 
					\State $\mathbb{D} \leftarrow \mathbb{D} \cup c_j$
				\EndIf
			\EndFor
		\EndFor

		\ForAll{$d_i \in \mathbb{D}$}
			\State $t_{new} \leftarrow$ \textsc{Initialise($d_i$)}
			\State $\mathbb{T} \leftarrow \mathbb{T} \cup t_{new}$
			\State $\mathbb{U} \leftarrow \mathbb{U} \cup t_{new}$
		\EndFor

		\ForAll{$t_i \in \mathbb{T} \setminus \mathbb{U}$}
			\State $S_{avg_i} \leftarrow$ \textsc{Update($W$)}
		\EndFor

		\ForAll{$t_i \in \mathbb{T}$}
			\If{$S_{avg_i} > \epsilon_{accept}$ \textbf{and} $d_i > N_{minRead}$}
				\State $t_i$ is marked as a \emph{pallet}
            		\Else{ \textbf{if} $S_{avg_i} < \epsilon_{reject}$ \textbf{or} $t_i$ has not been detected for more than $\tau_{timeout}$}
                			\State $\mathbb{T} \leftarrow \mathbb{T} - t_i$
			\EndIf
		\EndFor
	\EndLoop
\EndFunction
\end{algorithmic}
\end{algorithm}

Online pallet tracking is described in Algorithm \ref{alg:trackingphase} with more detail.
The set of candidate pallets to track (i.e., the corresponding ROIs) is referred to as $\mathbb{T}$, is initially empty (line 3), and it is updated as long as the Algorithm proceeds.
The set $\mathbb{D}$ of \textit{unassigned} possible candidates and the set $\mathbb{U}$ of \textit{updated} pallet candidates are initialised (lines 4 and 5), and updated afterwards.
At each iteration, the Algorithm first calls the \textsc{Acquisition} routine (line 7), and the set $\mathbb{C}$ of candidate pallets is determined.
Then, for each already tracked candidate pallet $t_i \in \mathbb{T}$, a number of parameters are retrieved, i.e., the number of times $d_i$ it has been detected (line 9), its average confidence $S_{avg_i}$ (line 10), and its pose $p_i$ (line 11).
Afterwards, its predicted pose is computed using the robot velocity $V$ (line 12).

For each candidate pallet $c_j$ in the current acquisition, one of the following cases is foreseen:
\begin{itemize}
\item if the associated ROI closely matches with the expected pose $p_i$ of an already tracked candidate pallet $t_i$ (line 16), the ROI is associated with the same candidate, the candidate pallet's pose $p_i$ is updated with the new observation (line 17), the number of times $d_i$ the pallet has been detected is increased (line 18), the confidence $S_{avg_i}$ in the candidate is updated taking the average of each detection's confidence score in a recent time window $W$ (line 19), and $t_i$ is labelled as updated (line 20); data association is achieved by computing how much of the two relative bounding boxes overlap and comparing the result to an acceptance threshold $\epsilon_{overlap}$;
\item if the candidate pallet $c_j$ does not match with sufficient precision any currently tracked candidate, it starts to be tracked as a new candidate, and it is labelled as unassigned (line 22).
\end{itemize}
For all unassigned candidate pallets, a corresponding tracked candidate pallet $t_{new}$ is generated and initialised (lines 27-29).
If a tracked candidate pallet does not match any detected prospect ROI (line 25), then it is assumed as currently not visible and the average confidence $S_{avg_i}$ in that candidate pallet decreases (line 32).
The Algorithm can attempt to take a decision on every currently tracked candidate $t_i$ based on the associated confidence:
\begin{itemize}
\item if the average confidence $S_{avg_i}$ exceeds a given threshold $\epsilon_{accept}$, and it has been detected for more than $N_{minRead}$ times (line 35), then the candidate is recognised as a pallet;
\item if the average confidence $S_{avg_i}$ decreases below a threshold $\epsilon_{reject}$, or the candidate has not been detected for a number of times (line 37), it is removed.
\end{itemize}
Also note that more than a candidate can be confirmed at any time, effectively allowing to track multiple pallets in the environment, if present.
Finally, the detection, localisation, and tracking system loops through these steps, and whenever a detection is confirmed, it communicates to the robot control architecture the pallet's pose $p_i$, so that further action can take place (e.g., approaching the pallet).

Considering trade-offs, such an approach adds a small delay from the instant a pallet is first detected by the classifier to the moment when it is actually recognized as such by the system, allowing the robot to act on the pallet. On the other side though, we argue that such delay is usually very short even on modest hardware, and can be managed acting on the choice of parameters used in Algorithm \ref{alg:trackingphase}, especially $\epsilon_{accept}$ and $N_{minRead}$. Considering also the moderate speed at which these robots are meant to operate, this is a reasonable trade-off in order to achieve extremely few false positives and a more stable detection of true positives. 

\section{Experimental Validation}
\label{sec:experimental_results}

\subsection{Setup and Implementation}
\label{sec:system_overview}

Our setup employs a commercially available 2D laser rangefinder from SICK AG, model S3000 Pro CMS.
The rangefinder is connected to a PC through a RS422-USB converter. 
\textcolor{red}{As we described in Section \ref{sec:reference_scenario}, the forklift is assumed to mount two units of such sensor, but only one will be employed for pallet detection and tracking.}
The sensor has a maximum range of $49$ $m$ ($20$ $m$ at $20\%$ reflectivity), a resolution of $0.25$ $deg$, a maximum $16$ $Hz$ refresh frequency, and an empirical error of $0.003$ $m$. 
The maximum field of view of the rangefinder is $190$ $deg$, which is largely sufficient for the detection of objects in front of the robot.
As mentioned in Section \ref{sec:system_architecture}, the sensor generates an array of points expressed in polar coordinates, each array having size $761$.

Two different computers have been used for experimental validation: a lower-end PC is used online for range data acquisition and pallet detection, whereas a more powerful workstation has been adopted for offline training and testing the proposed networks.
In particular, the former is equipped with an Intel\textsuperscript{\textregistered} Core i5-4210U $1.70$ $GHz$ CPU and $6$ GB of RAM, and runs Ubuntu 16.04 $64$ bit.
The latter mounts an Intel\textsuperscript{\textregistered} Core i7-4790 $3.60$ $GHz$ CPU, $32$ GB of RAM and an Nvidia Geforce\textsuperscript{\textregistered} GTX970 GPU, and runs Ubuntu 14.04 $64$ bit.

The overall architecture has been implemented using MATLAB \textcolor{red}{2017b} and ROS-based software components. 
In particular, range data are acquired using an \emph{ad hoc} ROS node developed in C++, whereas pallet detection, classification and tracking are implemented in MATLAB using the Computer Vision System Toolbox.
The Robotics System Toolbox has been used to interface MATLAB with ROS in order to perform online pallet detection, classification and tracking, and offline training of the two neural networks. 

\textcolor{red}{We decided to run our algorithm at a lower target frequency than the maximum one allowed by the sensor, namely $4$ $Hz$, by feeding to our system an average of every four consecutive frames. In the coming sections, we will show that the proposed solution is able to achieve performance far beyond the requirements described in Section \ref{sec:reference_scenario}, despite such limitation in operating frequency and the current implementation being an unoptimized proof of concept.}



\subsection{Offline Training}
\label{sec:training}

The first step in our experiments consisted in training the neural networks described in the previous Section.
A dataset has been collected in an indoor environment (measuring 40 {m$^2$}), including pallets, trolleys, multiple obstacles such as walls as well as other robots, and furniture.
The dataset consists of $340$ 2D range scans, each one corresponding to a frame from the 2D laser rangefinder located in a different position, as presented in details in \cite{mohamed20192d}.
Raw data are converted to 2D bitmap-like images and augmented by generating new \textit{artificial} images, obtained by rotating the original images clockwise and anticlockwise of $90$ $deg$.
As a consequence, the total dataset consists of $1020$ images.
Each image is resized to $250 \times 250$ and stored in a CSV file.
In the file, each line corresponds to a single 2D bitmap-like image which has mainly two entries for training the Faster R-CNN detector. The first entry is the path to each image, while the second entry contains the ROI labels in the image, i.e., pallet.
More different entries are added, after the Faster R-CNN processing step, so as to list all the ROIs detected in the image in two classes for training the CNN-based classifier.

\begin{figure}
\renewcommand{\figurename}{Fig.}
\centering
\subfigure[Before suppression.]{
\label{fig:suppression}
\includegraphics[width=.35\columnwidth]{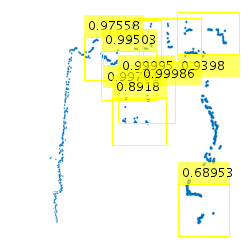}
}~
\subfigure[After suppression.]{
\label{fig:aftersupp}
\includegraphics[width=.35\columnwidth]{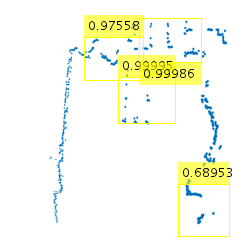}
}
\caption{On the left hand side, the ROIs provided by the Faster R-CNN and their confidence scores. On the right hand side it is shown how ROIs are reduced using non-maximum suppression with an overlap threshold of $0.3$.}
\end{figure}
In order to train the Faster R-CNN detector, the whole dataset has been divided in two parts: $70\%$ ($714$ samples) as a training set and $30\%$ ($306$ samples) as a test set.
Stochastic Gradient Descent (SGD) has been used to train the network and the initial learning rate $\alpha$ has been set to $\num{e-6}$.
The training process runs for $20$ epochs, leading to an approximately $45$-minute training time on our workstation.
Once training is complete and all the ROIs are generated, the corresponding bounding boxes are additionally filtered using non-maximum suppression with an overlap threshold of $0.3$, as shown in \eqref{eq:ovalue}.
Figure \ref{fig:suppression} shows a sample image, the ROIs detected in it, and their corresponding confidence scores, while Figure \ref{fig:aftersupp} shows only the ROIs remaining after suppression.

\begin{figure}[ht!]
\renewcommand{\figurename}{Fig.}
\begin{center}
\includegraphics[width=\columnwidth]{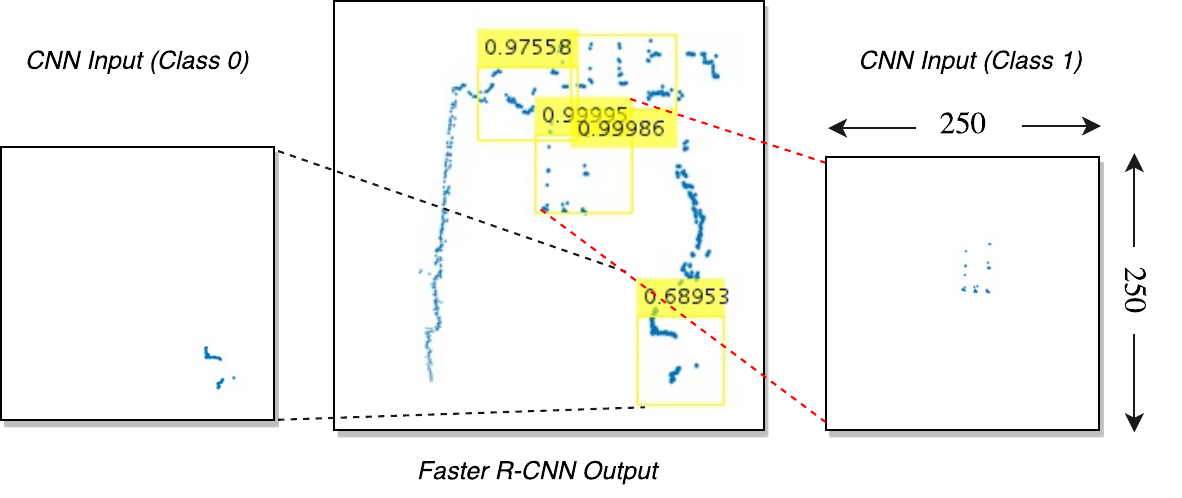}
\end{center}
\caption{An example of how training data for the CNN-based classifier are generated from the ROIs obtained by the Faster R-CNN detector. From each ROI a new image is generated, having the same size of the original one but only containing the detected object. The image on the left hand side (\emph{Class 0}), represents a generic object, while the one on the right hand side (\emph{Class 1}), represents a pallet.}
\label{fig:inputcnn}
\end{figure}
As anticipated above, the set of all ROIs obtained through this procedure can be labelled and used as an input for training the CNN-based classifier.
Considering for example the case in Figure \ref{fig:aftersupp}, $4$ different ROIs are detected and therefore $4$ new images are generated with the same size as the original one, but including only data inside the ROI's bounding box.
This process is better depicted in Figure \ref{fig:inputcnn}.
\emph{Class 0} objects (i.e., objects unlikely to be pallets) are sorted out by \emph{Class 1} objects (i.e., pallets) on the basis of the confidence score associated with the related bounding box, as shown in \eqref{eq:sequential}.
Considering that each image has to be labelled manually, a smaller set is actually used to train the CNN-based classifier, i.e., only an amount of images that are strictly necessary to achieve satisfying accuracy results on the test set are used.
Hence, $950$ images have been randomly selected and labelled among the available samples: $450$ of them represent a pallet (\emph{Class 1} in Figure \ref{fig:inputcnn}), while the other $500$ represent some undefined object (\emph{Class 0} in Figure \ref{fig:inputcnn}).
SGD and \emph{$k$-fold cross-validation} (with $k=10$) are used to train the CNN-based classifier with an initial learning rate $\alpha = 0.03$, and mini-batch size set to $50$, leading to an $99.58\%$ accuracy on the test set after a $26$ minutes training time. 

\subsection{Online Validation of the Pallet Tracking Process}
After having developed the architecture described in Section \ref{sec:pallettracking}, and having trained the Faster R-CNN detector and the CNN-based classifier, we have run several real-world experiments to validate the proposed approach.
\textcolor{red}{The first step was to gather the data generated by the laser rangefinder sensor as it moved with constant velocity in our test area. As previously stated in Section \ref{sec:system_overview}, this has been performed at an effective frequency of $4$ $Hz$, while the sensor has been manually moved at around $0.5$ $m/s$, which is much faster than the reference scenario requirement of $0.2$ $m/s$.
Twelve different trajectories about $4$ $m$ long have been recorded, each involving $40$ scans. The different trajectories are grouped in $4$ sets, which differ by the initial and final rangefinder's position, the pallet's pose in terms of position and orientation, and the location, size and shape of obstacles in the area. Trajectories in the same set differs by the path taken while approaching the pallet, i.e., with sensor approaching directly the pallet or keeping the target either on the right or left side.
The recorded trajectories also take into account dynamic obstacles, such as humans or other piece of equipement that could enter or leave the scene.}

In this Section, we report our analysis of the employed pallet tracking approach.
As far as the used parameters are concerned, the time window size $W$ on which we compute the average confidence score on a tracked candidate pallet has been set to $6$, whereas the maximum time $\tau_{timeout}$ a tracked candidate pallet can be invisible before being discarded by the system has been set to $0.6$. The average confidence acceptance $\epsilon_{accept}$ and the average confidence rejection $\epsilon_{reject}$ thresholds have been set to $0.6$ and $0.35$, respectively, whereas the minimum number of times (i.e., frames) $N_{minRead}$ a tracked candidate pallet must be acquired before it can be confirmed as a pallet given a sufficient average confidence score is set to $10$.

\begin{figure*}
\renewcommand{\figurename}{Fig.}
\centering
\begin{overpic}[width=0.32\textwidth,tics=10]{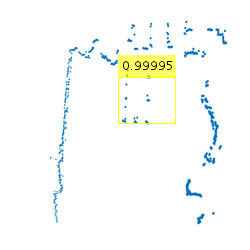}
    \put (5,85) {frame 1}
\end{overpic}
\begin{overpic}[width=0.32\textwidth,tics=10]{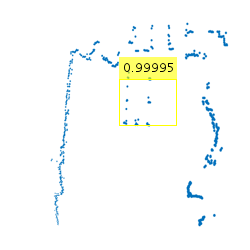}
    \put (5,85) {frame 2}
\end{overpic}
\begin{overpic}[width=0.32\textwidth,tics=10]{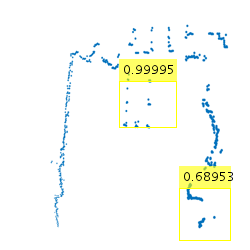}
    \put (5,85) {frame 3}
\end{overpic}
\begin{overpic}[width=0.32\textwidth,tics=10]{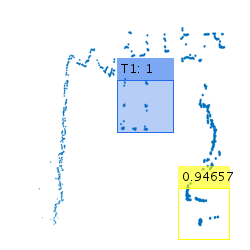}
    \put (5,85) {frame 6}
\end{overpic}
\begin{overpic}[width=0.32\textwidth,tics=10]{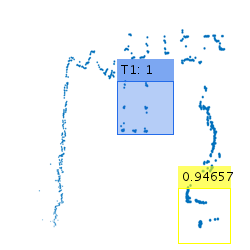}
    \put (5,85) {frame 7}
\end{overpic}
\begin{overpic}[width=0.32\textwidth,tics=10]{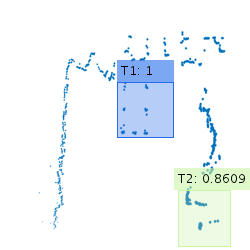}
    \put (5,85) {frame 8}
\end{overpic}
\begin{overpic}[width=0.32\textwidth,tics=10]{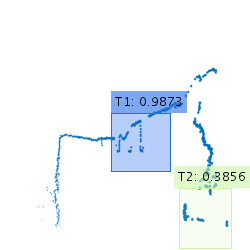}
    \put (5,85) {frame 15}
\end{overpic}
\begin{overpic}[width=0.32\textwidth,tics=10]{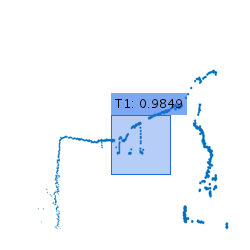}
    \put (5,85) {frame 16}
\end{overpic}
\begin{overpic}[width=0.32\textwidth,tics=10]{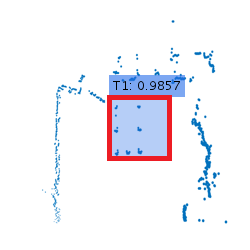}
    \put (5,85) {frame 17}
\end{overpic}

\caption{An example of single pallet tracking. Each image represents a frame. Yellow: tracked ROIs. Filled boxes: ROIs with a stable tracking. Red: ROIs  confirmed to be pallets. See text for details.}
\label{fig:OnePalletTracking}
\end{figure*}

{\textit{Single pallet tracking}}.
We applied the process illustrated in Algorithm \ref{alg:trackingphase} to real-world data obtained by imposing the trajectories described above.
For all of the twelve trajectories, the approach is able to detect the pallet and avoid false positives.
As an example, Figure \ref{fig:OnePalletTracking} shows the salient events in one of these trajectories up to frame $17$.
In the frames, each ROI represents a tracked candidate pallet.
For the sake of clarity, ROIs are generically contoured in yellow when they first appear in the robot's field of view, and then get assigned with an identification number and characteristic colour only after they are detected more than $N_{minRead}/2$ times and in case they have a high confidence score.
In frame 1, a ROI is immediately detected (T1), while a second one appears in frame $3$ (T2). 
T2 is a weak candidate, and its average confidence becomes less than $\epsilon_{reject} = 0.35$ in frame 16.
Consequently, the Algorithm stops tracking that candidate considering it as a likely false positive (i.e., a \emph{Class 0} detection).
It is noteworthy that higher values for $\epsilon_{reject}$ lead the Algorithm to delete weak candidates faster, but also increase the likelihood of deleting a true positive pallet.
As a consequence, $\epsilon_{reject}$ should be kept fairly small.
On the other hand, in frame 17 the Algorithm takes a positive decision on T1 as its average confidence surpasses $\epsilon_{accept}$ and it has been detected at least $N_{minRead}$ times.

\textcolor{red}{The success of our tests shows that the proposed architecture allows pallet detection and tracking at a speed $150\%$ higher than the one specified in our reference scenario requirements, despite running at a lower frequency than the maximum one allowed by the sensor. In details, the average computation time for a frame is $0.1358$ $s$ with a variance of $\num{1.6853e-4}$ $s^2$.
This result suggests that it is possible to run our pallet tracking algorithm at a frequency of almost $8$ $Hz$ on the current hardware, a $100\%$ higher frequency than the one we decided to adopt. On this side, significant performance gains may be achieved through system optimizations, such as a native C++ implementation instead of a MATLAB-based one, making use of GPU computing as it is increasingly common in autonomous vehicles, or just employ a more powerful machine rather than the mid-range  one we employed here for testing purposes. This way, it may be possible to push the system up to the sensor-imposed $16$ $Hz$ limit, allowing the algorithm to be applied to vehicles moving much faster than what is currently common in internal logistics applications.}

\textcolor{red}{Finally, on a more qualititative side,} notice that due to the simple nature of the data provided by the 2D laser rangefinder, the system \textcolor{red}{could} be prone to false positives. Yet, we have not experienced any issues in our testing despite objects with similar contours were present in the environment. This may be partially due to the online pallet tracking strategy outlined in Section \ref{sec:pallettracking}.

\begin{figure*}
\renewcommand{\figurename}{Fig.}
\centering
\begin{overpic}[width=0.32\textwidth,tics=10]{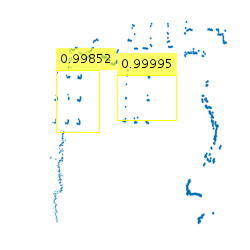}
    \put (5,85) {frame 1}
\end{overpic}
\begin{overpic}[width=0.32\textwidth,tics=10]{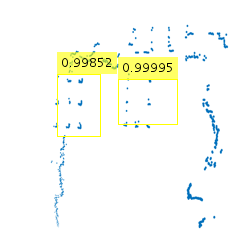}
    \put (5,85) {frame 3}
\end{overpic}
\begin{overpic}[width=0.32\textwidth,tics=10]{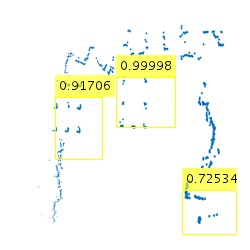}
    \put (5,85) {frame 5}
\end{overpic}
\begin{overpic}[width=0.32\textwidth,tics=10]{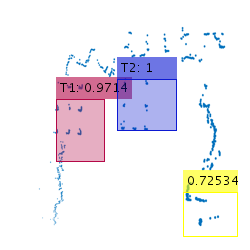}
    \put (5,85) {frame 6}
\end{overpic}
\begin{overpic}[width=0.32\textwidth,tics=10]{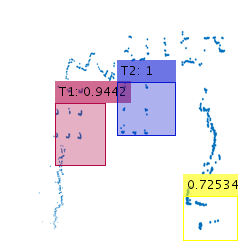}
    \put (5,85) {frame 8}
\end{overpic}
\begin{overpic}[width=0.32\textwidth,tics=10]{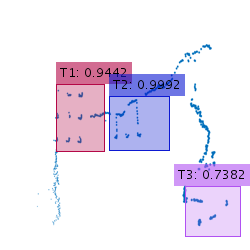}
    \put (5,85) {frame 10}
\end{overpic}
\begin{overpic}[width=0.32\textwidth,tics=10]{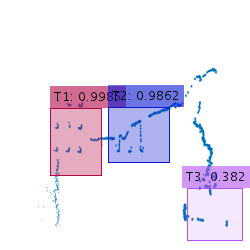}
    \put (5,85) {frame 15}
\end{overpic}
\begin{overpic}[width=0.32\textwidth,tics=10]{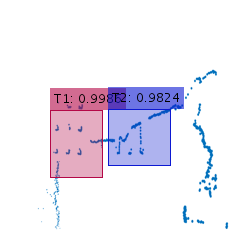}
    \put (5,85) {frame 16}
\end{overpic}
\begin{overpic}[width=0.32\textwidth,tics=10]{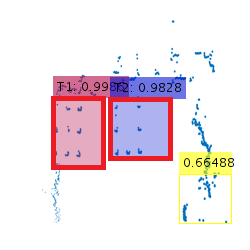}
    \put (5,85) {frame 17}
\end{overpic}

\caption{An example of multiple pallets tracking, same conventions as in Fig. \ref{fig:OnePalletTracking}. Regions of different colors indicate ROIs recognized simultaneously.}
\label{fig:TwoPalletsTracking}
\end{figure*}

\textit{Multiple pallets tracking}.
As anticipated above, we used artificially generated data to get preliminary results in scenarios possibly involving multiple pallets. This was not part of the reference scenario presented in Section \ref{sec:reference_scenario}, but it was worth exploring as it is an intrinsic property of the proposed system.
Our results confirm the ability to detect pallets and avoid false positives even in the case two pallets are present in front of the robot.
Figure \ref{fig:TwoPalletsTracking} depicts one example using the same graphical representation as the one introduced above.
In this case, the Algorithm detects three possible candidates in the first five frames, but the third one (T3) is dropped on frame 16, due to its low average confidence score. Furthermore, the first two (T1 and T2) are stronger candidates and they finally get accepted as pallets in frame 17. Concerning performance metrics, \textcolor{red}{we have not observed any significant loss compared to the single pallet tracking case}. 

At the moment, the main limitation in the multiple pallet case is that the system is not able to univocally identify pallets, but only distinguish them with respect to each other. This was not a limitation in our reference scenario as described in Section \ref{sec:reference_scenario}, but it can be an issue in other applications. We will explore in the future solutions to this problem, for example employing identification markers, which are much less prone to robustness issues compared to localization ones.

\section{Conclusions}
The paper presents and discusses a possible \textcolor{red}{application of object detection with CNNs to the problem of detecting, localising, and tracking pallets using 2D laser rangefinder data only.}
This is achieved by converting 2D rangefinder data into bitmap-like images where CNNs can look for possible candidate pallets.
Pallet candidates detected by the two CNNs \textit{in cascade} are passed down to a Kalman filter based tracker, which allows for having an estimate of pallet positions at any time, even when they are momentarily not visible, as well as helping the system filter out false positives.
In the paper, we detail the proposed architecture and present detection and classification results.
We conclude that our approach is a viable solution to correctly detect, localise and track pallets reliably, while attaining reasonable performance for real-world applications. 

Future work includes refining the precision of position estimate for pallets with respect to a robot-centred reference frame, as well as integrating orientation estimation and running a series of experiments onsite to further validate the approach. \textcolor{red}{Such experiments will also allow us to provide more data on the correlation between the algorithm operating frequency and maximum speed the vehicle can keep for robust tracking, and optimize our implementation accordingly for applications and scenarios that require stronger performances.}
On the functionality side, we intend to explore the possibility of targeting pallet types or different objects at the same time, in so far as the approach offers a certain degree of modularity, since it is possible to add more CNN-based classifiers in parallel without the necessity to retrain the existing networks or substantially modify the tracking algorithm.
Finally, we will explore methods that may allow in the the future to univocally identify an item identified and tracked by the system.


\bibliographystyle{spmpsci}
\bibliography{references}

\begin{thebibliography}{10}
\providecommand{\url}[1]{{#1}}
\providecommand{\urlprefix}{URL }
\expandafter\ifx\csname urlstyle\endcsname\relax
  \providecommand{\doi}[1]{DOI~\discretionary{}{}{}#1}\else
  \providecommand{\doi}{DOI~\discretionary{}{}{}\begingroup
  \urlstyle{rm}\Url}\fi

\bibitem{aref2013position}
Aref, M.M., Ghabcheloo, R., Kolu, A., Hyvonen, M., Huhtala, K., Mattila, J.:
  Position-based visual servoing for pallet picking by an
  articulated-frame-steering hydraulic mobile machine.
\newblock In: Proceedings of the 2013 IEEE Conference on Robotics, Automation
  and Mechatronics (RAM). Manila, Philippines (2013)

\bibitem{aref2016multistage}
Aref, M.M., Ghabcheloo, R., Kolu, A., Mattila, J.: A multistage controller with
  smooth switching for autonomous pallet picking.
\newblock In: Proceedings of th 2016 IEEE International Conference on Robotics
  and Automation (ICRA). Stockholm, Sweden (2016)

\bibitem{aref2014macro}
Aref, M.M., Ghabcheloo, R., Mattila, J.: A macro-micro controller for pallet
  picking by an articulated-frame-steering hydraulic mobile machine.
\newblock In: Proceedings of the 2014 IEEE International Conference on Robotics
  and Automation (ICRA). Hong Kong, China (2014)

\bibitem{arlot2010survey}
Arlot, S., Celisse, A.: A survey of cross-validation procedures for model
  selection.
\newblock Statistics surveys \textbf{4}, 40--79 (2010)

\bibitem{asvadi2017depthcn}
Asvadi, A., Garrote, L., Premebida, C., Peixoto, P., Nunes, U.J.: Depthcn:
  Vehicle detection using {3D-LIDAR and ConvNet}.
\newblock In: Intelligent Transportation Systems (ITSC), 2017 IEEE 20th
  International Conference on, pp. 1--6. IEEE (2017)

\bibitem{baglivo2008object}
Baglivo, L., Bellomo, N., Miori, G., Marcuzzi, E., Pertile, M., De~Cecco, M.:
  An object localization and reaching method for wheeled mobile robots using
  laser rangefinder.
\newblock In: Proceedings of the 2008 International IEEE Conference Intelligent
  Systems (IS). Varna, Bulgaria (2008)

\bibitem{baglivo2011autonomous}
Baglivo, L., Biasi, N., Biral, F., Bellomo, N., Bertolazzi, E., Da~Lio, M.,
  De~Cecco, M.: Autonomous pallet localization and picking for industrial
  forklifts: a robust range and look method.
\newblock Measurement Science and Technology \textbf{22}(8), 085,502 (2011)

\bibitem{beder2007comparison}
Beder, C., Bartczak, B., Koch, R.: A comparison of {PMD}-cameras and
  stereo-vision for the task of surface reconstruction using patchlets.
\newblock In: Proceedings of the 2007 IEEE Conference on Computer Vision and
  Pattern Recognition (CVPR). Minneapolis, MN, USA (2007)

\bibitem{bostelman2006visualization}
Bostelman, R., Hong, T., Chang, T.: Visualization of pallets.
\newblock In: Proceedings of SPIE, the International Society for Optical
  Engineering. Boston, MA, USA (2006)

\bibitem{brust2015convolutional}
Brust, C.A., Sickert, S., Simon, M., Rodner, E., Denzler, J.: Convolutional
  patch networks with spatial prior for road detection and urban scene
  understanding.
\newblock arXiv preprint arXiv:1502.06344  (2015)

\bibitem{byun2008real}
Byun, S., Kim, M.: Real-time positioning and orienting of pallets based on
  monocular vision.
\newblock In: Proceedings of the 2008 IEEE International Conference on Tools
  with Artificial Intelligence (ICTAI). Daytona, OH, USA (2008)

\bibitem{Capezioetal2011}
Capezio, F., Mastrogiovanni, F., Scalmato, A., Sgorbissa, A., Vernazza, P.,
  Vernazza, T., Zaccaria, R.: Mobile robots in hospital environments: an
  installation case study.
\newblock In: Proceedings of the 2011 European Conference on Mobile Robotics
  (ECMR). Örebro, Sweden (2011)

\bibitem{chen2012pallet}
Chen, G., Peng, R., Wang, Z., Zhao, W.: Pallet recognition and localization
  method for vision guided forklift.
\newblock In: Proceedings of the 2012 International Conference on Wireless
  Communications, Networking and Mobile Computing (WiCOM). Shanghai, China
  (2012)

\bibitem{Cucchiara2000FocusBF}
Cucchiara, R., Piccardi, M., Prati, A.: Focus based feature extraction for
  pallets recognition.
\newblock In: Proceedings of the 11th British Machine Vision Conference (BMVC).
  Bristol, UK (2000)

\bibitem{cuevas2005kalman}
Cuevas, E.V., Zaldivar, D., Rojas, R.: Kalman filter for vision tracking.
\newblock Technical Report, Freie Universität Berlin, Inst. Informatik,
  Berlin, Germany  (2005)

\bibitem{cui2010robust}
Cui, G.Z., Lu, L.S., He, Z.D., Yao, L.N., Yang, C.X., Huang, B.Y., Hu, Z.H.: A
  robust autonomous mobile forklift pallet recognition.
\newblock In: Proceedings of the 2010 International Asia Conference on
  Informatics in Control, Automation and Robotics (CAR). Wuhan, China (2010)

\bibitem{DAndrea2012}
D'Andrea, R.: A revolution in the warehouse: a retrospective on {K}iva systems
  and the grand challenges ahead.
\newblock IEEE Transactions on Automation Science and Engineering
  \textbf{4}(9), 638--639 (2012)

\bibitem{Darvishetal2018}
Darvish, K., Wanderlingh, F., Bruno, B., Simetti, E., Mastrogiovanni, F.,
  Casalino, G.: Flexible human-robot cooperation models for assisted shop-floor
  tasks.
\newblock arXiv preprint arXiv:1707.02591  (2018)

\bibitem{feng2018towards}
Feng, D., Rosenbaum, L., Dietmayer, K.: Towards safe autonomous driving:
  Capture uncertainty in the deep neural network for lidar {3D} vehicle
  detection.
\newblock arXiv preprint arXiv:1804.05132  (2018)

\bibitem{garibott1996robolift}
Garibotto, G., Masciangelo, S., Ilic, M., Bassino, P.: Robolift: a vision
  guided autonomous fork-lift for pallet handling.
\newblock In: Proceedings of the 1996 IEEE/RSJ International Conference On
  Intelligent Robots And Systems (IROS). Osaka, Japan (1996)

\bibitem{garibotto1997service}
Garibotto, G., Masciangelo, S., Ilic, M., Bassino, P.: Service robotics in
  logistic automation: Robolift: vision based autonomous navigation of a
  conventional fork-lift for pallet handling.
\newblock In: Proceedings of the 1997 International Conference on Advanced
  Robotics (ICAR). Monterey, CA, USA (1997)

\bibitem{Ghosh1970}
Ghosh, B.: Sequential tests of statistical hypotheses.
\newblock Addison-Wesley (1970)

\bibitem{girshick2015fast}
Girshick, R.: Fast {R-CNN}.
\newblock In: Proceedings of the 2015 IEEE International Conference on Computer
  Vision (ICCV). Santiago, Chile (2015)

\bibitem{girshick2014rich}
Girshick, R., Donahue, J., Darrell, T., Malik, J.: Rich feature hierarchies for
  accurate object detection and semantic segmentation.
\newblock In: Proceedings of the 2014 IEEE Conference on Computer Vision and
  Pattern Recognition (CVPR). Washington, DC, USA (2014)

\bibitem{he2010feature}
He, Z., Wang, X., Liu, J., Sun, J., Cui, G.: Feature-to-feature based laser
  scan matching for pallet recognition.
\newblock In: Proceedings of the 2010 International Conference on Measuring
  Technology and Mechatronics Automation (ICMTMA). Changsha City, China (2010)

\bibitem{hebert1986outdoor}
Hebert, M.: Outdoor scene analysis using range data.
\newblock In: Proceedings of the 1986 IEEE International Conference on Robotics
  and Automation (ICRA). San Francisco, CA, USA (1986)

\bibitem{Heyer2010}
Heyer, C.: Human-robot interaction and future industrial robotics applications.
\newblock In: Proceedings of the 2010 IEEE/RSJ International Conference on
  Intelligent Robots and Systems (IROS). Taipei, Taiwan (2010)

\bibitem{hoffman1987segmentation}
Hoffman, R., Jain, A.K.: Segmentation and classification of range images.
\newblock IEEE Transactions on Pattern Analysis and Machine Intelligence (5),
  608--620 (1987)

\bibitem{holz2016fast}
Holz, D., Behnke, S.: Fast edge-based detection and localization of transport
  boxes and pallets in {RGB-D} images for mobile robot bin picking.
\newblock In: Proceedings of the 2016 International Symposium on Robotics
  (ISR). Munich, Germany (2016)

\bibitem{kim2001model}
Kim, W., Helmick, D., Kelly, A.: Model based object pose refinement for
  terrestrial and space autonomy.
\newblock In: Proceedings of the 6th International Symposium on Artificial
  Intelligence, Robotics and Automation in Space. Montreal, Canada (2001)

\bibitem{Krugeretal2009}
Krüger, J., Lien, T., Verl, A.: Cooperation of humans and machines in the
  assembly lines.
\newblock CIRP Annals - Manufacturing Technology \textbf{58}(2), 628--646
  (2009)

\bibitem{kuhn1955hungarian}
Kuhn, H.W.: The hungarian method for the assignment problem.
\newblock Naval Research Logistics (NRL) \textbf{2}(1-2), 83--97 (1955)

\bibitem{lecking2006variable}
Lecking, D., Wulf, O., Wagner, B.: Variable pallet pick-up for automatic guided
  vehicles in industrial environments.
\newblock In: Proceedings of the 2006 IEEE Conference on Emerging Technologies
  and Factory Automation (ETFA). Prague, Czech Republic (2006)

\bibitem{Lecun2015}
LeCun, Y., Bengio, Y.E., Hinton, G.: Deep learning.
\newblock Nature \textbf{521}, 436--444 (2015)

\bibitem{liang2018deep}
Liang, M., Yang, B., Wang, S., Urtasun, R.: Deep continuous fusion for
  multi-sensor {3D} object detection.
\newblock In: Proceedings of the European Conference on Computer Vision (ECCV),
  pp. 641--656 (2018)

\bibitem{Mastrogiovannietal2010}
Mastrogiovanni, F., Sgorbissa, A., Zaccaria, R.: From autonomous robots to
  artificial ecosystems.
\newblock In: H.~Nakashima, H.~Aghajan, J.C. Augusto (eds.) Handbook of Ambient
  Intelligence and Smart Environments, pp. 635--668. Springer, Boston, MA, USA
  (2004)

\bibitem{Mastrogiovannietal2004}
Mastrogiovanni, F., Sgorbissa, A., Zaccaria, R.: A system for hierarchical
  planning in service mobile robotics.
\newblock In: Proceedings of the 8th International Conference on Intelligent
  Autonomous Systems (IAS-8). Amsterdam, The Netherlands (2004)

\bibitem{Mastrogiovannietal2007}
Mastrogiovanni, F., Sgorbissa, A., Zaccaria, R.: The more the better? a discuss
  about line features for self-localisation.
\newblock In: Proceedings of the 2007 IEEE/RSJ International Conference on
  Intelligent Robots and Systems (IROS). San Diego, CA, USA (2007)

\bibitem{Mastrogiovannietal2008}
Mastrogiovanni, F., Sgorbissa, A., Zaccaria, R.: Learning to extract line
  features: beyond split \& merge.
\newblock In: Proceedings of the 2008 International Conference on Intelligent
  Autonomous Systems (IAS-10). Baden-Baden, Germany (2008)

\bibitem{Mastrogiovannietal2009a}
Mastrogiovanni, F., Sgorbissa, A., Zaccaria, R.: Context assessment strategies
  for ubiquitous robots.
\newblock In: Proceedings of the 2009 IEEE International Conference on Robotics
  and Automation (ICRA). Atlanta, GA, USA (2009)

\bibitem{Mastrogiovannietal2009b}
Mastrogiovanni, F., Sgorbissa, A., Zaccaria, R.: Robust navigation in an
  unknown environment with minimal sensing and representation.
\newblock IEEE Transactions on Systems, Man, and Cybernetics, Part B:
  Cybernetics \textbf{39}(1), 212--229 (2009)

\bibitem{Mastrogiovannietal2013a}
Mastrogiovanni, F., Sgorbissa, A., Zaccaria, R.: How the location of the range
  sensor affects {EKF}-based localisation.
\newblock Journal of Intelligent \& Robotic Systems \textbf{68}(2), 121--145
  (2013)

\bibitem{matti2017combining}
Matti, D., Ekenel, H.K., Thiran, J.P.: Combining {LiDAR} space clustering and
  convolutional neural networks for pedestrian detection.
\newblock arXiv preprint arXiv:1710.06160  (2017)

\bibitem{mohamed2017detection}
Mohamed, I.S.: Detection and tracking of pallets using a laser rangefinder and
  machine learning techniques.
\newblock Master's thesis, European Master on Advanced Robotics Plus (EMARO+),
  University of Genova (2017)

\bibitem{mohamed20192d}
Mohamed, I.S., Capitanelli, A., Mastrogiovanni, F., Rovetta, S., Zaccaria, R.:
  A {2D} laser rangefinder scans dataset of standard {EUR} pallets.
\newblock Data in Brief p. 103837 (2019)

\bibitem{newman1993model}
Newman, T.S., Flynn, P.J., Jain, A.K.: Model-based classification of quadric
  surfaces.
\newblock CVGIP: Image Understanding \textbf{58}(2), 235--249 (1993)

\bibitem{Novikov2001}
Nov\'ikov, A.: Uniform asymptotic expansion of likelihood ratio for markov
  dependent observations.
\newblock Annals of the Institute of Statistical Mathematics \textbf{53}(4),
  799--809 (2001)

\bibitem{nygards2000docking}
Nygards, J., Hogstrom, T., Wernersson, A.: Docking to pallets with feedback
  from a sheet-of-light range camera.
\newblock In: Proceedings of the 2000 IEEE/RSJ International Conference On
  Intelligent Robots And Systems (IROS). Takamatsu, Japan (2000)

\bibitem{oh2014development}
Oh, J.Y., Choi, H.S., Jung, S.H., Kim, H.S., Shin, H.Y.: Development of pallet
  recognition system using {K}kinect camera.
\newblock International Journal of Multimedia and Ubiquitous Engineering
  \textbf{9}(4), 227--232 (2014)

\bibitem{pages2001computer}
Pages, J., Armangu{\'e}, X., Salvi, J., Freixenet, J., Mart{\'\i}, J.: A
  computer vision system for autonomous forklift vehicles in industrial
  environments.
\newblock In: Proceedings of the 2001 Mediterranean Conference on Control and
  Automation (MEDS). Dubrovnik, Croatia (2001)

\bibitem{pfister2003weighted}
Pfister, S.T., Roumeliotis, S.I., Burdick, J.W.: Weighted line fitting
  algorithms for mobile robot map building and efficient data representation.
\newblock In: Proceedings of the 2003 IEEE International Conference on Robotics
  and Automation ICRA. Taipei, Taiwan (2003)

\bibitem{premebida2005segmentation}
Premebida, C., Nunes, U.: Segmentation and geometric primitives extraction from
  {2D} laser range data for mobile robot applications.
\newblock Robotica \textbf{2005}, 17--25 (2005)

\bibitem{redmon2016you}
Redmon, J., Divvala, S., Girshick, R., Farhadi, A.: You only look once:
  Unified, real-time object detection.
\newblock In: 2016 IEEE Conference on Computer Vision and Pattern Recognition
  (CVPR), pp. 779--788. Las Vegas, NV, USA (2016)

\bibitem{Ren2017}
Ren, S., He, K., Girshick, R., Sun, J.: Faster {R-CNN}: Towards real-time
  object detection with region proposal networks.
\newblock IEEE Transactions on Pattern Analysis and Machine Intelligence
  \textbf{39}(6), 1137--1149 (2017)

\bibitem{schulenburg2003self}
Schulenburg, E., Weigel, T., Kleiner, A.: Self-localization in dynamic
  environments based on laser and vision data.
\newblock In: Proceedings of the 2003 IEEE/RSJ International Conference on
  Intelligent Robots and Systems (IROS). Las Vegas, NV, USA (2003)

\bibitem{seelinger2005automatic}
Seelinger, M., Yoder, J.D.: Automatic pallet engagment by a vision guided
  forklift.
\newblock In: Proceedings of the 2005 IEEE International Conference on Robotics
  and Automation (ICRA). Barcelona, Spain (2005)

\bibitem{syu2016computer}
Syu, J.L., Li, H.T., Chiang, J.S., Hsia, C.H., Wu, P.H., Hsieh, C.F., Li, S.A.:
  A computer vision assisted system for autonomous forklift vehicles in real
  factory environment.
\newblock Multimedia Tools and Applications \textbf{76}(4), 18,387--18,407
  (2017)

\bibitem{teng2010real}
Teng, Z., Kim, J.H., Kang, D.J.: Real-time lane detection by using multiple
  cues.
\newblock In: Proceedings of the 2010 International Conference on Control
  Automation and Systems (ICCAS). Gyeonggi-do, South Korea (2010)

\bibitem{uijlings2013selective}
Uijlings, J.R., Van De~Sande, K.E., Gevers, T., Smeulders, A.W.: Selective
  search for object recognition.
\newblock International Journal of Computer Vision \textbf{104}(2), 154--171
  (2013)

\bibitem{varga2015improved}
Varga, R., Costea, A., Nedevschi, S.: Improved autonomous load handling with
  stereo cameras.
\newblock In: Proceedings of the 2015 IEEE International Conference on
  Intelligent Computer Communication and Processing (ICCP). Cluj-Napoca,
  Romania (2015)

\bibitem{varga2014vision}
Varga, R., Nedevschi, S.: Vision-based autonomous load handling for automated
  guided vehicles.
\newblock In: Proceedings of the 2014 IEEE International Conference on
  Intelligent Computer Communication and Processing (ICCP), pp. 239--244.
  Cluj-Napoca, Romania (2014)

\bibitem{varga2016robust}
Varga, R., Nedevschi, S.: Robust pallet detection for automated logistics
  operations.
\newblock In: Proceedings of the 11th Joint Conference on Computer Vision,
  Imaging and Computer Graphics Theory and Applications (VISIGRAPH) - Volume 4:
  VISAPP, pp. 470--477. Rome, Italy (2016)

\bibitem{Wald1945}
Wald, A.: Sequential tests of statistical hypotheses.
\newblock The Annals of Mathematical Statistics \textbf{16}(2), 117--186 (1945)

\bibitem{walter2015situationally}
Walter, M.R., Antone, M., Chuangsuwanich, E., Correa, A., Davis, R., Fletcher,
  L., Frazzoli, E., Friedman, Y., Glass, J., How, J.P.: A situationally aware
  voice-commandable robotic forklift working alongside people in unstructured
  outdoor environments.
\newblock Journal of Field Robotics \textbf{32}(4), 590--628 (2015)

\bibitem{walter2010closed}
Walter, M.R., Karaman, S., Frazzoli, E., Teller, S.: Closed-loop pallet
  manipulation in unstructured environments.
\newblock In: Proceedings of the 2010 IEEE/RSJ International Conference on
  Intelligent Robots and Systems (IROS), pp. 5119--5126. Taipei, Taiwan (2010)

\bibitem{wang2016autonomous}
Wang, S., Ye, A., Guo, H., Gu, J., Wang, X., Yuan, K.: Autonomous pallet
  localization and picking for industrial forklifts based on the line
  structured light.
\newblock In: Proceedings of the 2016 IEEE International Conference on
  Mechatronics and Automation (ICMA). Harbin, China (2016)

\bibitem{weichert2013automated}
Weichert, F., Skibinski, S., Stenzel, J., Prasse, C., Kamagaew, A., Rudak, B.,
  Ten~Hompel, M.: Automated detection of euro pallet loads by interpreting pmd
  camera depth images.
\newblock Logistics Research \textbf{6}(2-3), 99--118 (2013)

\bibitem{zhou2017voxelnet}
Zhou, Y., Tuzel, O.: Voxelnet: End-to-end learning for point cloud based {3D}
  object detection.
\newblock arXiv preprint arXiv:1711.06396  (2017)

\end{thebibliography}

\end{document}